\documentclass[11pt, a4paper, logo, copyright]{eai}
\usepackage{shlab}

\usepackage[authoryear, sort&compress, round]{natbib}
\usepackage[]{mdframed}
\usepackage[utf8]{inputenc}
\usepackage{dblfloatfix}

\usepackage{amsmath,amsfonts,bm}
\usepackage{multirow}
\usepackage{subcaption}
\usepackage{wrapfig}

\def\eqref#1{equation~\ref{#1}}

\def\1{\bm{1}}

\DeclareMathAlphabet{\mathsfit}{\encodingdefault}{\sfdefault}{m}{sl}
\SetMathAlphabet{\mathsfit}{bold}{\encodingdefault}{\sfdefault}{bx}{n}

\usepackage{listings}
\usepackage{hyperref}
\usepackage{url}
\usepackage{graphicx}
\usepackage{amsmath} %
\usepackage{mathrsfs} %
\usepackage{etoolbox}
\usepackage{cleveref}
\usepackage{tcolorbox}
\usepackage{xcolor}
\usepackage{colortbl}
\usepackage{booktabs}       %
\usepackage{amsfonts}       %
\usepackage{nicefrac}       %
\usepackage{microtype}      %
\usepackage{subcaption}
\usepackage{algorithm}
\usepackage{algorithmic}
\usepackage{multirow}
\usepackage{subcaption}
\usepackage{lipsum}
\usepackage{amssymb}   
\usepackage{xspace}

\usepackage{pifont} 
\usepackage{xcolor} 
\usepackage{adjustbox} 

\usepackage{array}
\usepackage{adjustbox}
\usepackage{makecell}

\usepackage{enumitem}%
\setlist[itemize]{noitemsep, topsep=0pt}
\usepackage{enumitem,kantlipsum}

\newlength\savewidth

\definecolor{baselinecolor}{HTML}{d6eaf8}

\definecolor{mygray}{gray}{0.4}

\newcommand{\frameworkname}{{InternVLA-M1}}

\AtBeginEnvironment{tcolorbox}{\tiny}

\newcommand{\cmark}{\ding{51}}
\newcommand{\xmark}{\ding{55}}

\newcount\Comments  %
\usepackage{color}
\definecolor{darkred}{rgb}{0.9,0,0}
\definecolor{darkblue}{rgb}{0,0,0.7}
\definecolor{purple}{rgb}{.6, 0,.6}
\definecolor{orange}{rgb}{1.0,0.64,0}
\definecolor{green}{rgb}{0, 0.5, 0}
\newcommand{\kibitz}[2]{\ifnum\Comments=1\textcolor{#1}{#2}\fi}

\definecolor{darkergreen}{RGB}{19,168,33}

\title{\frameworkname: A Spatially Guided Vision-Language-Action Framework for Generalist Robot Policy}
\correspondingauthor{Contributors are listed in \Cref{sec:authors}.}

\renewcommand{\today}{2025-10-15}

\author{Intern Robotics, Shanghai AI Laboratory}

\begin{document}

\begin{abstract}
We introduce \frameworkname, a unified framework for spatial grounding and robot control that advances instruction-following robots toward scalable, general-purpose intelligence. Its core idea is \textbf{spatially guided vision-language-action training}, where spatial grounding serves as the critical link between instructions and robot actions. \frameworkname\ employs a two-stage pipeline: (i) spatial grounding pre-training on over 2.3M spatial reasoning data to determine ``where to act'' by aligning instructions with visual, embodiment-agnostic positions, and (ii) spatially guided action post-training to decide ``how to act'' by generating embodiment-aware actions through plug-and-play \textbf{spatial prompting}. This spatially guided training recipe yields consistent gains: InternVLA-M1 outperforms its variant without spatial guidance by +14.6\% on SimplerEnv Google Robot, +17\% on WidowX, and +4.3\% on LIBERO Franka, while demonstrating stronger spatial reasoning capability in box, point, and trace prediction. To further scale instruction following, we built a simulation engine to collect 244K generalizable pick-and-place episodes, enabling a 6.2\% average improvement across 200 tasks and 3K+ objects. In real-world clustered pick-and-place, InternVLA-M1 improved by 7.3\%, and with synthetic co-training, achieved +20.6\% on unseen objects and novel configurations. Moreover, in long-horizon reasoning-intensive scenarios, it surpassed existing works by over 10\%. These results highlight spatially guided training as a unifying principle for scalable and resilient generalist robots.

\links{
  \link{code}{Code:InternVLA-M1}{https://github.com/InternRobotics/InternVLA-M1}, 
  \link{model}{Model:InternVLA-M1}{https://huggingface.co/collections/InternRobotics/internvla-m1-68c96eaebcb5867786ee6cf3}, 
  \link{data}{Data:InternData-M1}{https://huggingface.co/datasets/InternRobotics/InternData-M1}, 
  \link{homepage}{Homepage}{https://internrobotics.github.io/internvla-m1.github.io/}
}

\end{abstract}

\maketitle


\section{Introduction}

Large multimodal foundation models~\cite{llavaov, chen2024internvl, bai2025qwen2, clip, siglip} have demonstrated strong generalization by leveraging web-scale vision–language alignment and instruction-following corpora. To extend these capabilities into the physical domain, robots must not only understand \emph{what} an instruction means but also determine \emph{where} and \emph{how} to act in the 3D world. This gap is fundamental. Textual abstractions capture spatial cues only indirectly, whereas real-world actions demand continuous, embodied interactions that are scarcely represented in the training data of vision–language models (VLMs). Teleoperated datasets~\cite{open_x_embodiment, bu2025agibot, wu2024robomind, khazatsky2024droid} provide valuable supervision; yet, their scale and diversity remain modest compared to large instruction-following corpora. In this context, an embodiment-agnostic spatial prior, which functions as a bridge between textual instructions and embodiment-specific motor commands, offers a promising foundation for scalable robot learning.

\begin{figure}[ht!]
\centering
\includegraphics[width=\textwidth]{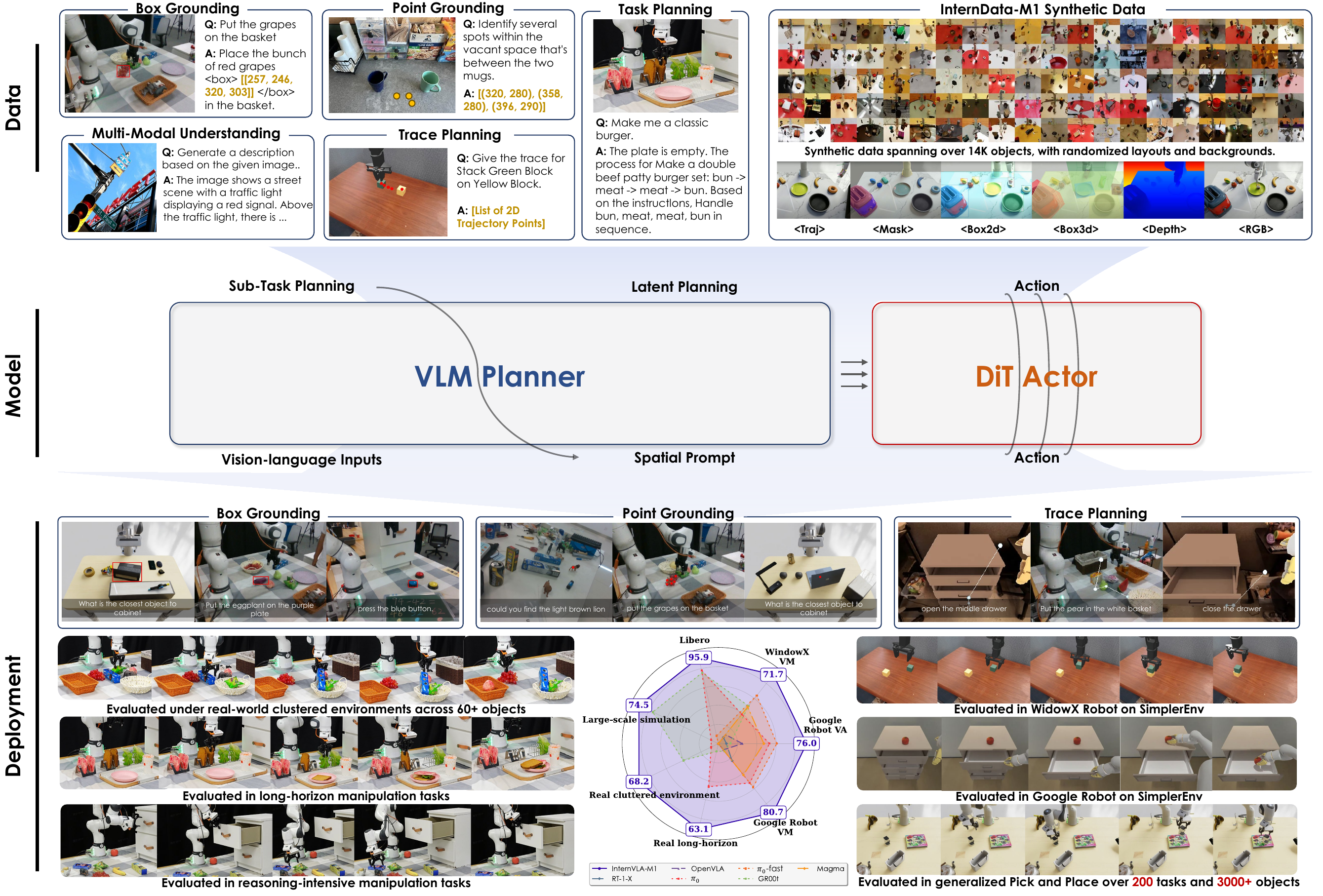}
\caption{InternVLA-M1 integrates spatial grounding into the vision–language–action training pipeline. Given a task instruction, the VLM planner produces latent plans through explicit spatial prompting, which then effectively guides the action expert to generate control signals.}
\label{fig:teaser}
\end{figure}

Prior work has approached this challenge through hierarchical robotic systems~\cite{voxposer, copa, moka, rekep, qi2025sofar, cao2025robobrain2, yuan2024robopoint}, which explicitly encode spatial priors using foundation models~\cite{anygrasp, sam, oquab2023dinov2} but often rely on rule-based task decomposition and manually designed planning heuristics. This rigid separation between symbolic task structures and low-level motor control makes such systems difficult to scale automatically to more complex and diverse tasks, particularly hindering end-to-end policy learning. In contrast, recent data-driven VLAs~\cite{openvla, RT-2, pi_0, hirobot, helix, molmoact} leverage pretrained vision-language models and large-scale teleoperation datasets~\cite{open_x_embodiment, khazatsky2024droid, bu2025agibot, wu2024robomind} to directly learn robot control. However, these models tend to overfit fine-grained motor behaviors while under-generalizing to high-level linguistic instructions that involve absolute or relational positions, thereby failing to fully incorporate spatial priors into execution. Core spatial priors such as object recognition, affordance grounding, visual trajectory reasoning, relative localization, and scaling provide transferable knowledge across robotic platforms. Once these priors are established, embodiment-specific learning can focus on concrete control strategies (e.g., manipulator joints, end-effector trajectories, humanoid locomotion, or mobile navigation). Such a division clarifies the role of spatial priors as general-purpose foundations while leaving embodiment-specific details to downstream adaptation, thereby bridging the gap between abstract instruction following and grounded physical execution.

Building on the separation between spatial priors and embodiment-specific control, we introduce {\frameworkname}, a dual-system vision–language–action framework that unifies high-level reasoning with grounded execution. {\frameworkname} consists of a spatial prior VLM planner that interprets linguistic instructions and reasons about spatial relations, and an action expert that translates these grounded representations into executable motor commands. To achieve this, we construct over 3M multimodal training samples, including 2.3M spatial grounding data and 0.7M multimodal understanding data collected from web, real-world, and simulated sources. To leverage the above pre-training data, we propose spatially guided two-stage training recipes: (i) \textit{spatial grounding pre-training} for the VLM, which establishes transferable spatial understanding through large-scale multimodal supervision on points, boxes, and traces; and (ii) \textit{spatially guided action post-training}, which specializes these priors for embodiment-specific control under joint supervision. This design bridges abstract goal reasoning with concrete physical execution, enabling robust instruction following across diverse and complex environments.

To validate these capabilities, we conduct a comprehensive evaluation across multiple benchmarks in both simulated environments and real-world settings. {\frameworkname} demonstrates strong generalization and robust performance across diverse scenarios:
\begin{itemize}[leftmargin=0.15in]
\item
On SimplerEnv (Google Robot and WidowX), InternVLA-M1 achieves a new state of the art, surpassing its variant by improving the average success rate by up to +5.9\% and +9.8\%, respectively. It also demonstrates strong spatial reasoning capabilities across box, point, and trace prediction tasks. Further analysis shows that spatially guided action post-training effectively transfers spatial reasoning ability to motor control.

\item For the generalizable pick-and-place 200 tabletop scenarios, our model exhibits strong generalization to unseen objects and instructions under few-demonstration fine-tuning, achieving an average improvement of 6.2\% over prior works.

\item In real-world settings, InternVLA-M1 demonstrates strong instruction-following capability, achieving a +20.6\% success rate on unseen objects and novel setups in clustered pick-and-place tasks. It also maintains robust long-horizon performance under perturbations (e.g., physical interference, task replanning), outperforming baselines such as GR00T and $\pi_0$ by large margins.

\end{itemize}

\section{\frameworkname}
\label{sec:method}

We propose {\frameworkname}, a dual-system, end-to-end vision–language–action (VLA) framework. 
It integrates both a language head and an action head within a single model (\Cref{sec: vla model}). 
The language head establishes instruction-to-visual grounding through spatial pretraining and co-training, 
while the action head conditions on these learned spatial priors to generate embodiment-specific motor commands(\Cref{sec:vlm_pre_training}). 
This joint design bridges abstract linguistic goals with grounded execution, enabling robust instruction following across diverse and complex scenes.
 
\subsection{Model Architecture}
\label{sec: vla model}

\begin{figure}[ht!]
    \centering
    \includegraphics[width=1.\textwidth]{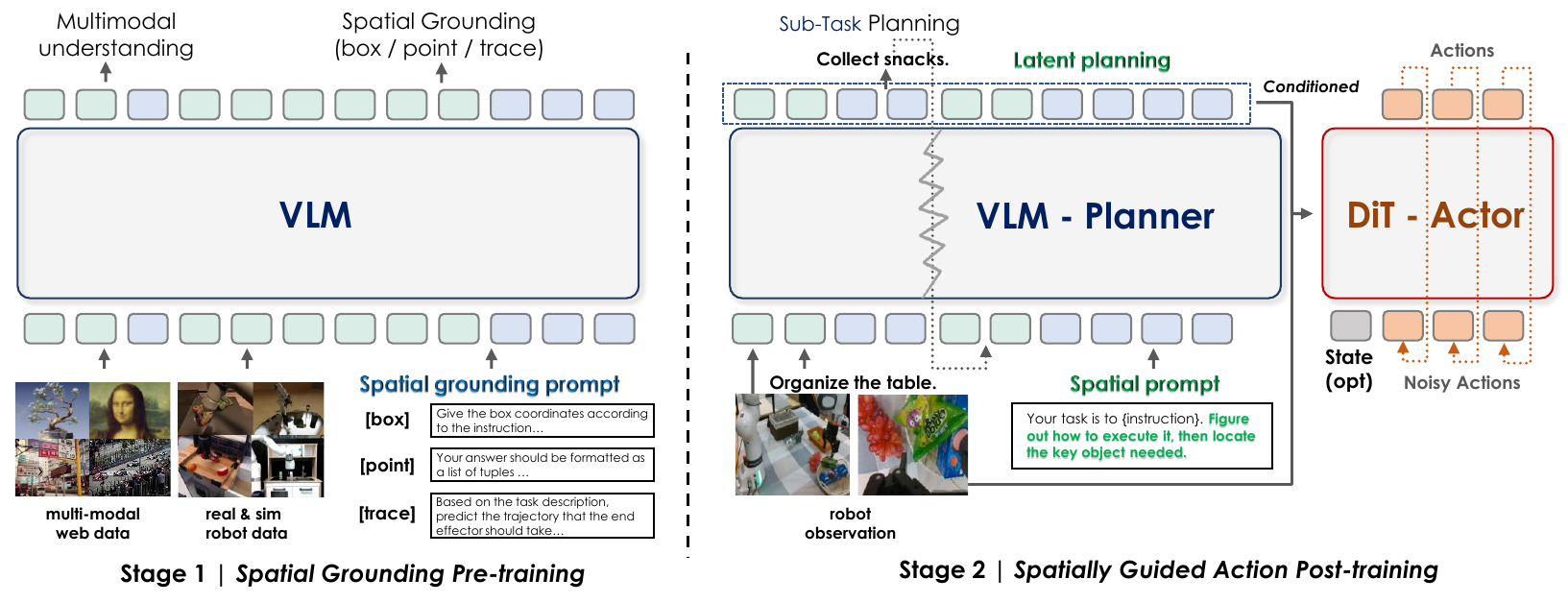}
    \caption{\textbf{Overview of \frameworkname.} \frameworkname\ adopts a spatially guided two-stage training pipeline. Stage 1 (spatial grounding pre-training): the VLM is trained on large-scale multisource multimodal spatial grounding data to learn embodiment-agnostic spatial priors. Stage 2 (spatially guided action post-training): the VLM Planner, functioning as a slow but reliable System 2 reasoner, generates latent planning tokens via spatial prompting as the condition to the action expert (instantiated as a DiT Actor) to execute as a fast System 1 controller.} 
    \label{fig:framework}
\end{figure}

\noindent\textbf{Dual-System.} 
\frameworkname\ is a dual-system, end-to-end VLA framework pre-trained on large-scale spatial grounding data collected from diverse sources. 
\frameworkname\ employs the Qwen2.5-VL-3B-instruct~\cite{qwen25vl} as the multimodal encoder for System 2, which is to capture spatial priors.
It adopts the diffusion policy~\cite{chi2023diffusionpolicy} (86 M) as the Action Expert (System 1, the fast executor), which effectively models embodiment-specific control. This expert is built on the DINOv2 visual encoder~\cite{oquab2023dinov2} (21 M) and a lightweight state encoder (0.4 M), forming a compact vision–action model.  In total, {\frameworkname} comprises approximately 4.1B parameters. During inference, the system runs on a single RTX 4090 GPU with around 12 GB of memory usage. With FlashAttention, the VLM component achieves inference speeds of approximately 10 FPS. Action execution can be further accelerated via chunking and KV caching. 

\noindent\textbf{Dual-Supervision.}  
The dual-system architecture supports both multimodal supervision and action supervision during training. 
In each training step, batches from both data types are jointly processed, and the model computes losses from the two supervision signals. 
The resulting gradients are aggregated and applied in a single optimization update, ensuring that perception and control are co-adapted rather than learned in isolation.  Specifically, the VLM planner is aligned with a broad range of spatial grounding data, both real and synthetic, covering tasks such as object detection, affordance recognition, and visual trajectory planning.
In parallel, the Action Expert is trained on robot demonstration data, enabling it to specialize these priors into embodiment-specific motor commands. 
This dual-supervision strategy establishes a cohesive link between high-level semantic perception and low-level motion control, which is essential for robust instruction following in both simulation and real-world settings.

\noindent\textbf{Latent planning via spatial prompting.} 
To connect the VLM Planner with the action expert, we adopt a lightweight querying transformer (8.7 MB) conditioned on the latent planning embeddings produced by the VLM Planner. The querying transformer stabilizes expert learning and inference by mapping variable-length input tokens into a fixed set of learnable query tokens. It is implemented as a $k$-layer cross-attention module, where the query tokens selectively attend to $k$ intermediate layers of the VLM (e.g., $k=1$ attends only to the final layer).

To explicitly activate the spatial perception capability learned during spatial grounding pre-training, we employ spatial prompting. For instance, in general object manipulation tasks, we append simple prompts such as “Figure out how to execute it, then locate the key object needed.” after the task instruction. The extracted feature embeddings provide the planner with explicit spatial cues that facilitate more reliable grounding. Motivated by prior studies~\cite{pi_05KI, chatvla2, bjorck2025gr00t} showing that direct gradient flow between action and VLM modules may distort multimodal knowledge, we introduce a gradient decay factor within the querying transformer. This attenuates the gradients propagated from the Action Expert back to the VLM (e.g., by a factor of 0.5), thereby preserving the Planner’s semantic reasoning ability while still enabling effective joint optimization.

\subsection{Training Recipe}
\label{sec:vlm_pre_training}

To leverage spatial priors for stronger embodiment-specific control in instruction following, 
{\frameworkname} adopts a spatially guided two-stage training pipeline:

\noindent\textbf{Stage 1: Spatial grounding pre-training.}
As shown in~\Cref{fig:framework}, the first stage optimizes only the VLM. The objective is not generic vision–language pre-training, but stronger spatial reasoning and planning ability essential for robotics. We combine internet-scale multimodal corpora with robot-specific datasets such as RefCOCO, RoboRefIt~\cite{vlgrasp}, A0~\cite{xu2025a0}, MolmoAct~\cite{molmoact}, and Pixmo-Points~\cite{pixmo2024}. All robot datasets are reformatted into a unified QA-style structure covering bounding-box detection, trajectory prediction, affordance recognition, and chain-of-thought reasoning. Aligning them with web-scale data enables training under the same supervised fine-tuning framework as conventional VLMs.

\noindent\textbf{Stage 2: Spatially guided action post-training.}\label{sec:vla_post_training} 
In this stage, both the VLM and Action Expert are jointly optimized on demonstration data, ensuring semantic understanding and motion generation remain tightly integrated. Two strategies are employed:

\begin{itemize}[leftmargin=0.15in]
    \item \textbf{Spatial prompting.} Before predicting actions, we prepend a spatial cue to the task instruction to elicit structured reasoning about object relationships and task constraints. For example, the instruction “store all toys into the toy box” can be augmented with: “Identify all relevant toys and their spatial relationships to the container.” Although the VLM does not explicitly output a response to this auxiliary cue, its inclusion improves spatial awareness and generalization in manipulation tasks.
    \item \textbf{Co-training with spatial grounding data.} Training alternates between robot trajectory data and grounding data. For trajectory data, both the VLM backbone and the action Expert are optimized with an L2 loss between predicted and ground-truth noise. For spatial grounding data, only the VLM backbone is updated via next-token prediction. This co-training scheme reinforces spatial reasoning while supporting efficient end-to-end optimization.
\end{itemize}

\section{Data}
\label{sec:data_main}

This section introduces the datasets used in {\frameworkname}, covering pre-training, mid-training, and post-training stages. For VLM pre-training, we construct large-scale spatial grounding datasets with point, box, and trajectory annotations to enhance spatial perception and vision-language alignment. Mid-training employs synthetic manipulation data to bridge pre-training knowledge and robotic execution. Post-training uses both simulated and real-world instruction-following data, including large-scale tabletop tasks and real-robot demonstrations for long-horizon manipulation.

\subsection{Spatial Grounding Data for Pre-training}
\label{sec:pretrain_state_1_data}

\begin{figure}[ht!]
    \centering
    \includegraphics[width=1.\textwidth]{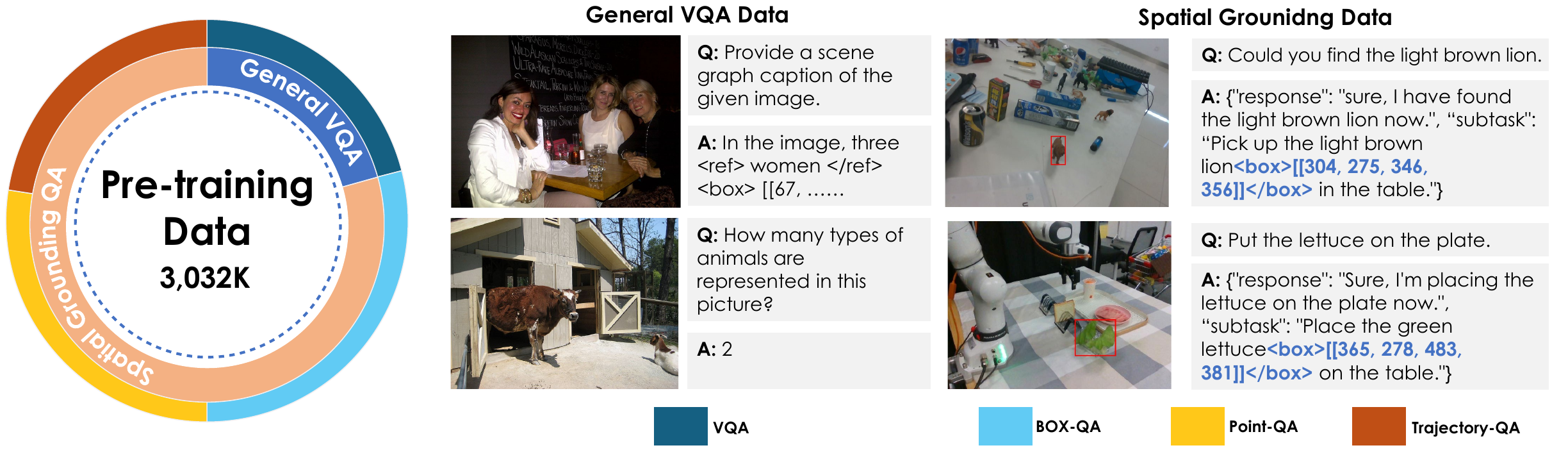}
    \caption{Overview of the pre-training data for the vision-language model. The data comprises two main parts: general VQA data to maintain the model's general multimodal capabilities, and spatial VQA data focusing on robotic-related grounding and spatial perception in a VQA format.}
    \label{fig: vlm pre-training data}
\end{figure}

The multimodal training dataset for our model comprises over 3M data, categorized into four distinct types: General QA, Box QA, Trajectory QA, and Point QA, as shown in \Cref{fig: vlm pre-training data}. Notably, more than 2.3M of these data are dedicated to spatial reasoning datasets. These categories ensure robust multimodal understanding while supporting adaptation to embodied tasks in tabletop robotic scenarios.  Below, we describe each category: 

\begin{itemize}[leftmargin=0.15in]
    \item \textbf{General QA}. Sourced from LLaVA-OneVision~\cite{li2024llava} and InternVL3~\cite{chen2024internvl, internvl3}, this category is sampled to cover diverse multimodal tasks, including image captioning, visual question answering (VQA), optical character recognition (OCR), knowledge grounding, and creative writing, resulting in approximately 637K samples in total.

    \item \textbf{Box QA}. We curate a diverse collection of multimodal grounding datasets, including RefCOCO~\cite{yu2016modeling, mao2016generation}, ASv2~\cite{wang2024all}, and COCO-ReM~\cite{singh2024benchmarking}, sourced from InternVL3~\cite{chen2024internvl, internvl3}. Additionally, we incorporate the InternData-M1 dataset, generated via scalable synthetic data generation as \Cref{sec: synthetic data engine}, and the RoboRefIt dataset~\cite{vlgrasp}, a specialized dataset for robotics grounding, resulting in approximately 879K samples in total.

    \item \textbf{Trajectory QA}. This category integrates the A0 ManiSkill subset~\cite{a0}, the InternData-M1 waypoint dataset, and the MolmoAct dataset~\cite{molmoact} to enable precise end-effector trajectory prediction. The A0 ManiSkill subset provides high-quality, object-centric trajectory data, where small objects move in coordination with the robotic arm's gripper. These trajectories can be approximated as end-effector movements for tabletop manipulation tasks, resulting in approximately 684K samples in total.
    
    \item \textbf{Point QA}. For precise point localization, we integrate multiple datasets, including the Pixmo-Points dataset~\cite{pixmo2024}, the RoboPoint dataset~\cite{yuan2024robopoint}, the RefSpatial dataset~\cite{zhou2025roborefer}, and a point subset extracted from the InternData-M1 dataset, each subjected to tailored preprocessing. Specifically, the Pixmo-Points dataset is filtered to exclude images with resolutions exceeding 1024 pixels and restricted to a maximum of 10 points per image. Additionally, we prioritize the extraction of object reference and region reference data from the RoboPoint and RefSpatial datasets to enhance grounding accuracy, resulting in approximately 832K samples in total.
\end{itemize}

All point coordinates are converted to absolute coordinates~\cite{qwen25vl}. Predicted coordinates are formatted in JSON and XML to support robust learning and adaptive processing of spatial instructions for diverse robotic tasks.

\subsection{Synthetic Data For Action Post-Pre-training}

To bridge the gap between VLM and VLA, we introduce a Post-Pre-Training phase, where large-scale simulated data is used to pre-train the VLA after VLM pre-training. This stage initializes the action head and facilitates the learning of action representations. Post-Pre-Training requires maintaining diversity both at the instruction and object levels. Consistent with the InternVLA-M1-Interface Data, we leverage GenManip as our data synthesis pipeline to construct a large-scale pick-and-place dataset, the InternData M1 dataset, which comprises 244K closed-loop samples. 
Specifically, we adopt the same object set and positional distributions as in InternVLA-M1-Interface Data, and process them through our scalable data pipeline. Each synthesized sample is rigorously validated to ensure correctness and consistency. To further enhance visual diversity, we introduce controlled randomization in lighting conditions and texture mappings. 

\subsection{Scalable Synthetic Data Engine for Instruction-Following}\label{sec: synthetic data engine}

\begin{figure}[t]
    \centering
    \includegraphics[width=1.\textwidth]{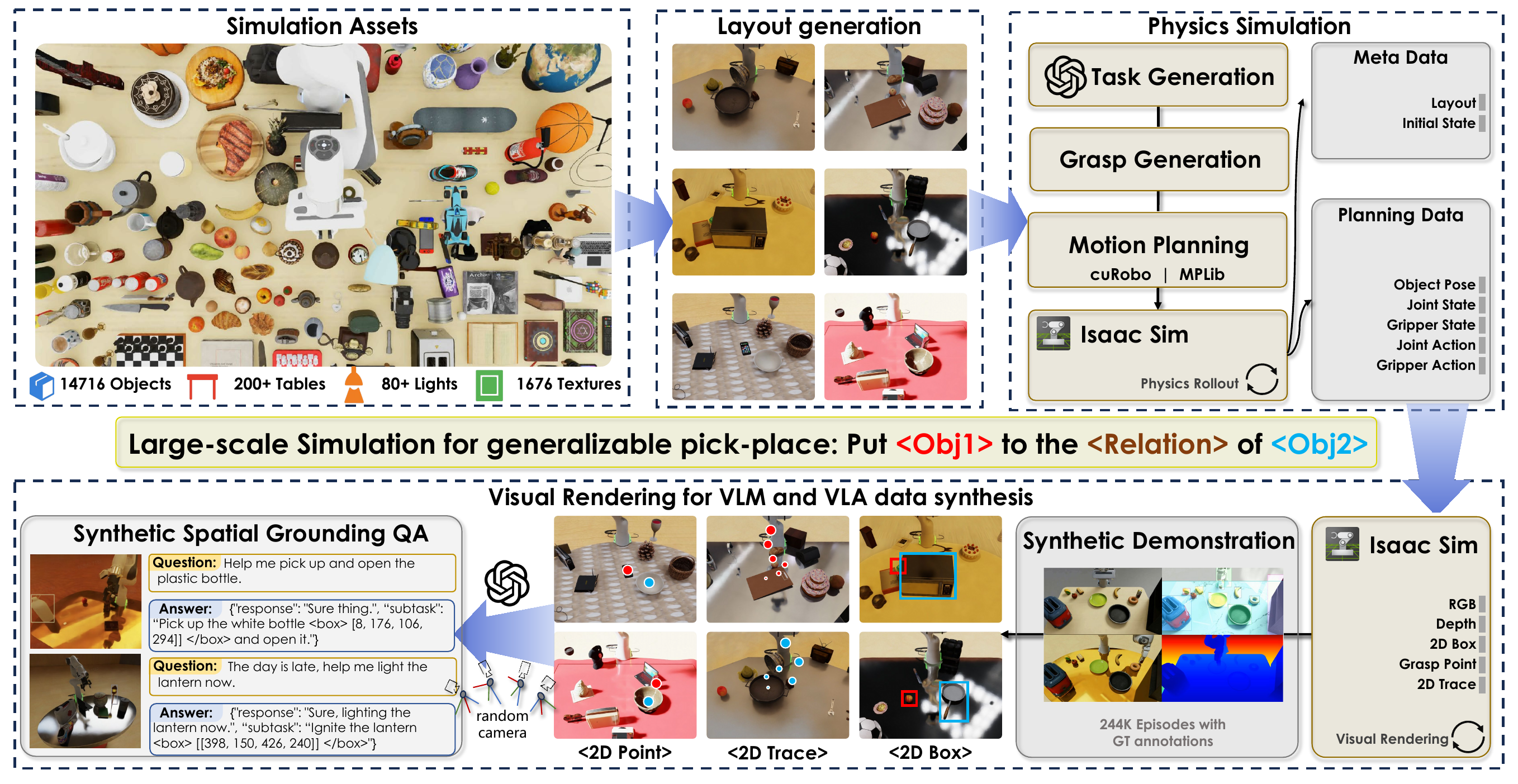}
    \caption{\textbf{Simulation data synthesis pipeline}. The pipeline generates diverse robotic manipulation data from a large asset library, converts intermediate representations into VQA data, and separates physics from rendering to reduce wasted failures and improve efficiency.}
    \label{fig:data-pipeline}
\end{figure}

To support large-scale end-to-end data generation for VLM pre-training, we build a highly scalable, flexible, and fully automated simulation pipeline on top of GenManip~\cite{gao2025genmanip} and Isaac Sim~\cite{isaac}. 

\noindent\textbf{Automatic task synthesis for generalizable pick-and-place.} 
We develop a scalable simulation pipeline (shown in~\Cref{fig:data-pipeline}) that generates diverse manipulation trajectories from randomized object layouts and lighting conditions. By leveraging privileged simulation signals including object poses, object meshes, and robot arm state, the system rapidly generates scene layouts via a scene graph solver and computes candidate grasps based on object meshes~\cite{liang2019pointnetgpd}. Each candidate trajectory is then executed once in physics for closed-loop verification, after which a scene-graph validator checks whether the task goals are achieved. Only trajectories that both execute successfully and pass validation are accepted, ensuring that all collected data are physically feasible and task-complete.

\noindent\textbf{Synthesis of VLM data and VLA data for spatial grounding.}
For higher efficiency, robot planning and rendering are fully decoupled in our framework. The planner records structured scene and trajectory data, including joint states, object positions, and action information, which are later replayed by the renderer under randomized lighting, materials, and viewpoints. To align the simulation with real world, we calibrate all cameras using ArUco markers, ensuring that their intrinsic and extrinsic parameters match those of real-world cameras, thus maintaining consistent viewpoint geometry. In addition to high-resolution images, the renderer produces rich intermediate outputs, such as object bounding boxes and 2D end-effector trajectories. These signals provide dense supervision for action learning and facilitate the creation of auxiliary datasets for tasks such as spatial grounding, affordance reasoning, and trajectory prediction. Our asset library includes 14K annotated objects, 211 tables, 1.6K textures, and 87 dome lights, offering data with high visual and physical diversity—critical for developing generalizable models.

\section{Experiments}
\label{sec:exp}

We conducted extensive experiments to evaluate the performance of {\frameworkname} in both simulation and real-world settings. First, we assess the performance on public simulated benchmarks (\Cref{exp:main_simpler}). Next, we fully evaluate the instruction-following of InternVLA-M1 for generalizable pick-and-place using Isaac-Sim (\Cref{exp:main_genman200}). Finally, we examine real-robot performance on long-horizon manipulation tasks to study instruction-following in real-world deployment (\Cref{sec:real_results}).

\vspace{-0.9 em}
\subsection{Experiments on Public Benchmarks}
\label{exp:main_simpler}

We use two established simulation suites:
\begin{itemize}[leftmargin=0.15in]
\item 
\textbf{SimplerEnv} is designed to probe robustness to visual appearance shifts. It includes both WidowX and Google Robot platforms, short-horizon atomic tasks, and controlled changes in lighting, color, surface texture, and camera pose. We report results on three task sets: Google Robot-VM (visual matching under viewpoint and lighting changes), Google Robot-VA (visual aggregation with varying textures and colors), and WidowX-VM (cross-robot generalization).

\item 
\textbf{LIBERO} is a language-conditioned manipulation suite built on a Franka arm with diverse scenes and expert demonstrations. We evaluate four task sets: LIBERO-Spatial (same objects, different spatial layouts), LIBERO-Object (fixed layout, different objects), LIBERO-Goal (fixed objects and layout, different goals), and LIBERO-Long (also known as LIBERO-10; longer tasks that span multiple objects, layouts, and operations).
\end{itemize}

\begin{table}[ht!]
  \centering
  \begin{adjustbox}{width=\linewidth}
  \begin{tabular}{ l l c c c c c c } 
    \toprule
    \makecell[c]{Google\\Robot} 
    & \multicolumn{1}{c}{Models} 
    & \multicolumn{1}{c}{Co-Train} 
    & \makecell[c]{Pick \\ Coke Can}
    & \makecell[c]{Move \\ Near} 
    & \makecell[c]{Open/Close \\ Drawer}
    & \makecell[c]{Open Top Drawer \\ and Place Apple} 
    & Avg \\
    \midrule
    \multirow{10}{*}{\makecell[l]{Visual\\Matching}}
      & RT-1~\cite{RT-1}     & \xmark & 85.7 & 44.2 & \underline{73.0} &  6.5 & 52.4 \\
      & RT-1-X~\cite{open_x_embodiment}  & \xmark& 56.7 & 31.7 & 59.7 & 21.3 & 42.4 \\
      & RT-2-X~\cite{RT-2} & \cmark & 78.7 & 77.9 & 25.0 &  3.7 & 46.3 \\
      & OpenVLA~\cite{openvla}  & \xmark  & 18.0 & 56.3 & 63.0 &  0.0 & 34.3 \\
      & {CogACT}~\cite{cogact}   & \xmark   & \underline{91.3} & \underline{85.0} & {71.8} & \underline{50.9} & 74.8 \\
       & SpatialVLA~\cite{spatialvla} & \xmark	& 86.0 &	77.9 & 	57.4 &	- & \underline{75.1} \\
      & $\pi_0$~\cite{pi_0}	& \xmark &  72.7  & 65.3 & 	38.3 & 	- &  58.8  \\
 & $\pi_0$-FAST~\cite{pertsch2025fast} & \xmark & 	75.3 & 	67.5 & 	42.9  & -  & 	61.9 \\
        & GR00T N1.5$^*$~\cite{bjorck2025gr00t} & \xmark & 51.7 & 54.0 & 27.8 & 7.4 & 35.2 \\ 
        & Magma~\cite{magma}  & \cmark & 83.7& 65.4  &  56.0 &  6.4 & 52.9 \\
        \cmidrule(lr){2-8}
    &   Vanilla VLA & \xmark   & {90.0} & {69.8} & {52.5} & {52.2} & {66.1} \\
    
    & \textbf{\frameworkname}      & \cmark &  \textbf{95.3} & \textbf{90.0} & \textbf{75.5} & \textbf{62.0} & \textbf{80.7} \\
    
      \rowcolor{gray!10}
      & $\Delta$  & & +5.3 & +20.2 & +23.0 & +9.8 & {\color{darkred}+14.6}  \\
      
    \midrule
    \multirow{10}{*}{\makecell[l]{Variant\\Aggregation}}
      & RT-1~\cite{RT-1}     & \xmark & \underline{89.8} & 50.0 & 32.3 &  2.6 & 43.7 \\
      & RT-1-X~\cite{open_x_embodiment} & \xmark & 49.0 & 32.3 & 29.4 & 10.1 & 30.2 \\
      & RT-2-X~\cite{RT-2} & \cmark & 82.3 & 79.2 & 35.3 & 20.6 & 54.4 \\
      & OpenVLA~\cite{openvla}  & \xmark  & 60.8 & 67.7 & 28.8 &  0.0 & 39.3 \\
      & {CogACT}~\cite{cogact}    & \xmark  & {89.6} & {80.8} & 28.3 & \underline{46.6} & {61.3} \\
       & SpatialVLA~\cite{spatialvla} & \xmark	& 88.0 &	\underline{82.5} & 	\underline{41.8} &	- & \underline{70.7} \\
        & $\pi_0$~\cite{pi_0}	& \xmark & 75.2 &	63.7 &	25.6 &	- & 54.8 \\
        & $\pi_0$-FAST~\cite{pertsch2025fast} & \xmark	& 77.6 &	68.2 &	31.3 & - &	59.0 \\
        & GR00T N1.5~\cite{bjorck2025gr00t} & \xmark & 69.3 & 68.7 & 35.8 & 4.0 & 44.5 \\
         & Magma~\cite{magma} & \cmark & 68.8 & 65.7  & 53.4 & 18.5 & 51.6 \\
        \cmidrule(lr){2-8}
        &   Vanilla VLA  & \xmark  & {92.3} & {80.3} & {50.1} & {31.4} & {63.5} \\
      
    & \textbf{\frameworkname}      & \cmark &  \textbf{86.1} & \textbf{82.0} & \textbf{72.0} & \textbf{64.0} & \textbf{76.0} \\
    
      \rowcolor{gray!10}
      & $\Delta$ &  & -6.2 & +1.7 & +21.9 & +32.6 & {\color{darkred}+12.5} \\
      \bottomrule
  \end{tabular}
  \end{adjustbox}
  \caption{Result comparisons of robotic manipulation on SimplerEnv (Google-Robot) benchmark. The underlined scores indicate the best results excluding {\frameworkname}. Numbers are officially reported; otherwise, we reimplement and mark such entries with $*$. }
  \label{tab:simpler-google} 
\end{table}

\noindent
\textbf{Baselines.}
We compare to state-of-the-art open VLA systems, including $\pi_{0}$~\cite{pi_0}, GR00T~\cite{bjorck2025gr00t}, OpenVLA~\cite{openvla}, CogACT~\cite{cogact}, and etc. We also include a {Vanilla VLA} built on \texttt{QwenVL-2.5-3B-Instruct} with a DiT action head. When available, we use official reported numbers; otherwise, we reimplement and mark such entries with $*$. We keep training data, observation spaces, and action type aligned with the most popular setups~\cite{cogact} to ensure a fair comparison.

\vspace{-0.8 em}
\subsubsection{SimplerEnv Benchmark}

\noindent\textbf{Experiment setup.} 
As described in \Cref{sec:vlm_pre_training}, we post-train {\frameworkname} on a subset of Open-X Embodiment (OXE) (including \texttt{fractal\_rt\_1} and \texttt{bridge\_v1}), with co-training on spatial grounding data (\Cref{sec:pretrain_state_1_data}).
The VLM takes the primary observation image, task instruction, and an auxiliary spatial prompt as input, while the action expert predicts actions with an action chunk size of 16. For multimodal data, the model follows an SFT-style question-answering format. Training is performed on 16 NVIDIA A100 GPUs for 50k steps (around 2.5 epochs), with total batch sizes of 256 for robot data and 64 for multimodal data, optimized with a summed loss over both data types. All evaluations are conducted within SimplerEnv using its official evaluation protocol. 

\begin{table}[ht!]
  \centering
  \begin{adjustbox}{width=\linewidth}
  \begin{tabular}{l l c c ccc c}
    \toprule
    \makecell[c]{WidowX\\Robot} 
    & \multicolumn{1}{c}{Models} 
    & \multicolumn{1}{c}{Co-Train} 
    & \makecell[c]{Put Spoon \\ on Towel} 
    & \makecell[c]{Put Carrot \\ on Plate} 
    & \makecell[c]{Stack Green Block \\ on Yellow Block} 
    & \makecell[c]{Put Eggplant \\ in Yellow Basket} 
    & Avg \\
    \midrule
    \multirow{9}{*}{\makecell[l]{Visual\\Matching}}
      & RT-1-X~\cite{RT-1} & \xmark & 0.0  & 4.2  & 0.0  & 0.0  & 1.1 \\
      & Octo-Base~\cite{octo}  & \xmark  & 15.8 & 12.5 & 0.0  & 41.7 & 17.5 \\
      & Octo-Small~\cite{octo} & \xmark  & 41.7 & 8.2  & 0.0  & 56.7 & 26.7 \\
      & OpenVLA~\cite{openvla}   & \xmark   & 4.2  & 0.0  & 0.0  & 12.5 & 4.2 \\
      & {CogACT}~\cite{cogact}      & \xmark        & {71.7}   & {50.8} & {15.0} & 67.5 & {51.3} \\
      & {SpatialVLA}~\cite{spatialvla}      & \xmark        & 16.7   & 25.0 & 29.2 & \underline{100.0} & 42.7 \\
    & {$\pi_0$}~\cite{pi_0}       & \xmark       & {29.1} & {0.0}  & {16.6} & {62.5} & 27.1 \\
    & $\pi_0$-FAST~\cite{pertsch2025fast} & \xmark 	& 29.1 & 21.9 & 10.8 & 66.6 & 48.3 \\
    & GR00T N1.5~\cite{bjorck2025gr00t} & \xmark  & \underline{75.3} & \underline{54.3} & \underline{57.0} & 61.3 & \underline{61.9} \\ 
    & Magma~\cite{magma} & \cmark  & 37.5 & 31.0 & 12.7  & 60.5 & 35.8 \\
    \cmidrule(lr){2-8}
      & Vanilla VLA & \xmark  & {56.6} & {63.3} & {27.0} & {71.8} & {54.7} \\
      
    & \textbf{\frameworkname}   & \cmark   & \textbf{87.5} & \textbf{67.9} & \textbf{31.3} & \textbf{100.0} & \textbf{71.7} \\
    
      \rowcolor{gray!10}
      & $\Delta$ & & +30.9 & +4.6 & +4.3 & +28.2 & \color{darkred}{+17.0} \\
      
    \bottomrule
  \end{tabular}
  \end{adjustbox}
  \caption{Result comparisons of robotic manipulation on SimplerEnv (WidowX) benchmark.
  The underlined scores indicate the best results excluding {\frameworkname}.
  }
  \label{tab:simpler-widowx}
\end{table}

\noindent\textbf{Result analysis.}
The main experimental results are presented in \Cref{tab:simpler-google} and \Cref{tab:simpler-widowx}. Compared with prior state-of-the-art models, it attains a 5.9\% gain in Google Robot Visual Matching, a 5.3\% gain in Visual Aggregation, and a 9.8\% gain on the WidowX benchmark. These results highlight the strong competitiveness of {\frameworkname} within the community. Compared to the Vanilla VLA based on Qwen2.5-VL-3B-Instruct, {\frameworkname} achieves substantial improvements: a 14.6\% increase in Google Robot Visual Matching and a 12.4\% increase in Visual Aggregation, along with a 17.0\% improvement on the WidowX benchmark. These results demonstrate the effectiveness of our spatially guided pre-training and action post-training strategies.

\begin{figure}[ht!]
    \centering
    \includegraphics[width=1.\textwidth]{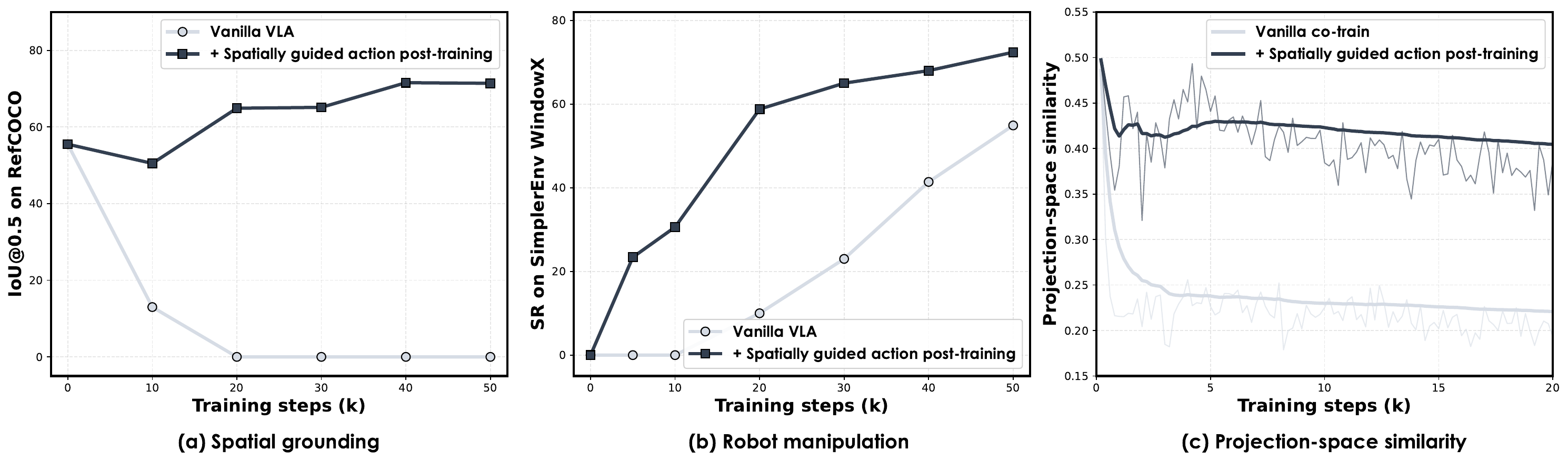}
     \caption{Ablation study on the effect of auxiliary spatial prompting for co-training robot manipulation with spatial grounding.
    From left to right: (a) spatial grounding performance (IoU@0.5 on RefCOCO-g);
    (b) manipulation performance (SimplerEnv-WidowX SR);
    (c) shows the gradient similarity of the spatial grounding and manipulation objectives.
    }
    \label{fig:abl_prompting}
\end{figure}

\noindent\textbf{Ablation study on dual-supervision co-training.}
\Cref{fig:abl_prompting} presents a comparative analysis of manipulation performance (WindowX) and perception performance (RefCOCO+) across training steps. The results demonstrate that omitting spatial data and spatially guided prompting during training leads to rapid degradation of spatial grounding capabilities and slower convergence in manipulation tasks. In contrast, \textit{spatially guided action post-training} accelerates convergence, substantially improves manipulation success rates, and enhances spatial grounding accuracy as shown in \Cref{fig:abl_prompting}.

To further analyze the relationship between the spatial grounding objective and the action manipulation objective, we compute the Projection-space Similarity (PSS)~\cite{raghu2017svcca} using Singular Value Decomposition (SVD). As shown in ~\Cref{fig:abl_prompting}(c), vanilla co-training of action data with spatial data yields a PSS of only 0.25, indicating significant misalignment between the gradient subspaces. In contrast, our spatially guided training approach increases the PSS to 0.42, demonstrating substantially improved optimization consistency. This enhanced alignment correlates with better preservation of spatial perception capabilities and faster convergence in manipulation tasks.

We conduct a comprehensive study of VLA training strategies and their effects across three distinct task categories: multi-modal understanding, spatial grounding, and robot manipulation performance, which is a type of generalist VLA. 
Specifically, we evaluate:
\begin{itemize}[leftmargin=0.15in]
\item {Multi-modal understanding:} MME~\cite{zhang2021mme}, MMVet~\cite{yu2023mm}, TextVQA~\cite{singh2019towards}, POPE~\cite{li2023evaluating}, COCO Caption~\cite{chen2015microsoft}
\item {Spatial grounding:} RefCOCO-g~\cite{mao2016generation} (Box IoU${0.5}$), Refit-testB~\cite{lu2023vl} (Box IoU${0.5}$), Where2Place~\cite{yuan2024robopoint} (evaluated by accuracy of predicted points with respect to ground-truth free space), and A0-maniskill~\cite{gu2023maniskill2} (evaluated using trajectory MAE, measuring mean absolute error between predicted and reference waypoints).
\item {Robot manipulation:} Google-Robot Visual Matching (VM), Variant Aggregations (VA), and WindowX Visual Matching (VM).
\end{itemize}

As shown in Table~\ref{tab:probing}, our {\frameworkname} achieves superior robotic manipulation performance while simultaneously preserving stronger multimodal understanding and spatial grounding capabilities compared to vanilla fine-tuning from VLM to VLA (Vanilla VLA) and direct co-training with spatial grounding data (vanilla co-train).

\begin{table}[ht!]
  \centering
  \caption{Ablation study of VLA training strategies across multi-modal understanding, spatial grounding, and robotic manipulation tasks.}
  \begin{adjustbox}{width=\linewidth}
  \begin{tabular}{l ccccc cccc ccc}
    \toprule

    & \multicolumn{5}{c}{Multi-modal Understanding}
    & \multicolumn{4}{c}{Spatial Grounding} 
    & \multicolumn{2}{c}{Robotic Manipulation} 
    \\
    \cmidrule(lr){2-6}
    \cmidrule(lr){7-10}
    \cmidrule(lr){11-12}
    
    \textbf{Models} 
      & \makecell[c]{MME} 
      & \makecell[c]{MMVet} 
      & \makecell[c]{TextVQA} 
      
      & \makecell[c]{POPE} 
      & \makecell[c]{COCO Caption \\ BLEU/ROUGE} 
      
      & \makecell[c]{RefCOCO-g \\ Box IoU$_{0.5}$$\uparrow$} 

      & \makecell[c]{Refit-testB \\ Box IoU$_{0.5}$$\uparrow$} 
    & \makecell[c]{Where2place \\ Point Acc$\uparrow$} 
      & \makecell[c]{A0-maniskill \\ Traj. MAE $\downarrow$} 
      & \makecell[c]{Google Robot \\ VM/VA} 
      & \makecell[c]{WindowX \\ VM} \\
    \midrule
    Vanilla VLA  & - & - & - & - & - & - & - & - & - & 66.1/63.5 & 54.7 \\
    Vanilla co-train & 1106 & 19.2 & 20.5 & 78.0 & 10.4/15.1 &  47.1  & 66.7 & 21.4 & 6.4 & 70.2/66.5 & 61.1 \\
    \midrule
    InternVLA-M1
    & 1411 & 23.3 & 28.6  & 86.2 & 13.0/13.4 & 71.2  & 74.3 & 25.5 & 5.1 & 80.7/76.0 & 71.7
    \\
    \bottomrule
  \end{tabular}
  \end{adjustbox}
  \label{tab:probing}
\end{table}

\subsubsection{LIBERO Benchmark}
\noindent\textbf{Experimental setups.}
Following \cite{openvla-oft}, we filter out failed demonstrations and pause frames. During training, the policy takes as input both wrist-mounted and third-person camera views. We fine-tune the model on each suite independently using 8 A100 GPUs with a batch size of 128 and an action chunk size of 8. Training runs for roughly 30K steps, lasting about 20 hours. Each suite is evaluated with 500 trials.

\begin{table}[ht!]
\small
\centering
\renewcommand{\arraystretch}{1.15}
\setlength{\tabcolsep}{6pt}

\begin{tabular}{l *{5}{c}}
\toprule
Models & Spatial & Objects & Goal & Long & Avg \\
\midrule
OpenVLA~\cite{openvla} & 84.7 & 88.4 & 79.2 & 53.7 & 76.5 \\
SpatialVLA~\cite{spatialvla} & 88.2 & 89.9 & 78.6 & 55.5 & 78.1 \\
CoT-VLA~\cite{zhao2025cot} & 87.5 & 91.6 & 87.6 & 69.0 & 83.9 \\
GR00T N1~\cite{bjorck2025gr00t}  & 94.4  & 97.6  & 93.0  & \underline{90.6} & 93.9  \\
$\pi_{0}$~\cite{pi_0}   & 96.8  & 98.8  & \underline{95.8}  & 85.2 & 94.2  \\
$\pi_{0}$-FAST~\cite{pertsch2025fast} & 96.4  & 96.8  & 88.6  & 60.2 & 85.5  \\
$\pi_{0.5}$-KI~\cite{pi_05KI} & \underline{98.0}  & \underline{97.8}  & 95.6  & 85.8 & \underline{94.3}  \\ \midrule
Vanilla VLA & \textbf{98.8} & 98.0 & 81.4 & 88.0 & 91.6 \\
\textbf{\frameworkname}  & 98.0 & \textbf{99.0} & \textbf{93.8} & \textbf{92.6} & \textbf{95.9}\\
\bottomrule
\end{tabular}
\caption{Result comparisons of robotic manipulation on LIBERO (Franka) benchmark.}
\label{tab:large-scale-simulated-experiments}
\end{table}

\noindent\textbf{Result analysis.}
The primary experimental results on the LIBERO benchmark are presented in \Cref{tab:large-scale-simulated-experiments}. Compared to previous strong baselines, such as GR00T N1 and $\pi_0$, the {\frameworkname} framework achieves notable improvements, particularly on the spatial and long-horizon tracks, with success rates of 98.0\% and 92.6\%, respectively. These results demonstrate the efficacy of our proposed method in managing complex, multi-step manipulation tasks. Specifically, for object placement, InternVLA-M1 attains a 99.0\% SR, which highlights its robust object grounding capability.

\subsection{Experiments on Instruction-Following in In-house Environment} 
\label{exp:main_genman200}

\subsubsection{Evaluation in Simulated Large-scale Pick-and-place}
Existing benchmarks such as SimplerEnv and LIBERO are limited in scale, which restricts the comprehensive evaluation of instruction-following manipulation in diverse and cluttered settings.
To more rigorously assess generalization capabilities, we conduct an experimental study on a large-scale simulation evaluation with enhanced object diversity and layout variation.

\noindent
\textbf{Experimental setups.}
We constructed 200 pick-and-place tasks based on Isaac-Sim~\cite{gao2025genmanip}, where the manipulated objects in each task are mutually distinct. Including background objects, the benchmark covers over 3K items and containers in total. Each task was executed once through the data generation pipeline to ensure its executability. Furthermore, for each of the 200 tasks, we additionally collected 5 trajectories with identical object sets but randomized layouts, which were used for post-training.
The observation space comprises two RGB images: one captured from a fixed third-person viewpoint and the other from a first-person camera mounted on the Franka end-effector. Both images are resized to $224 \times 224$ before being fed into the model. 
We fine-tune the model on each suite independently using 16 A100 GPUs, with a total batch size of 256 and an action chunk size of 16. Training is conducted for 20K steps.
Both our model and all baseline models are trained using delta joint space control.

\begin{figure}[ht!]
    \centering
    \includegraphics[width=1.\textwidth]{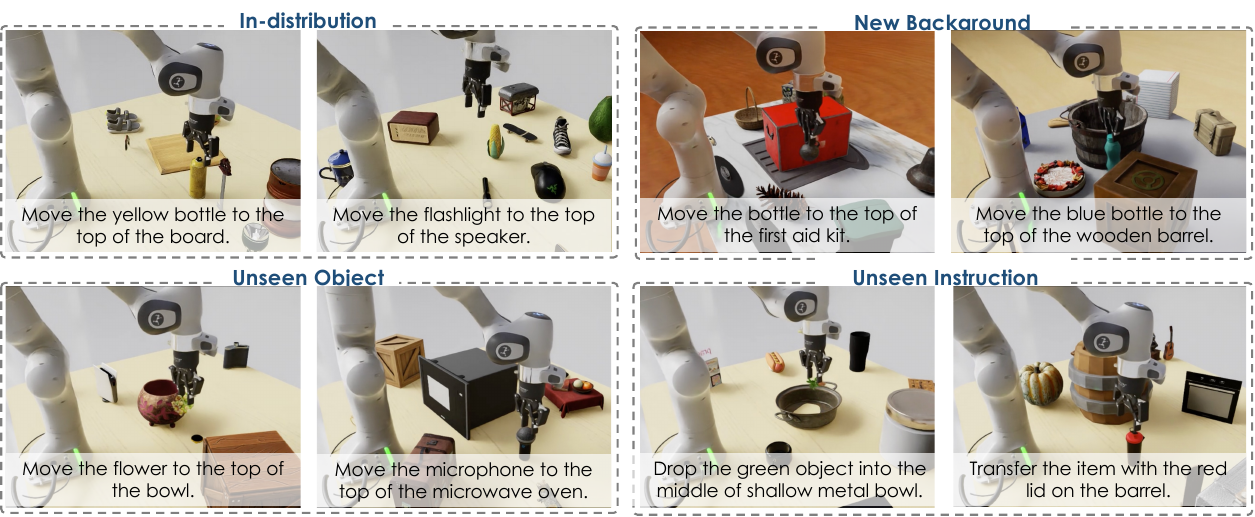}
    \caption{Evaluation settings for generalizable pick-and-place in large-scale simulation.}
    \label{fig:large_scale_simbench_200}
\end{figure}

\begin{figure}[ht!]
    \centering
    \includegraphics[width=1.\textwidth]{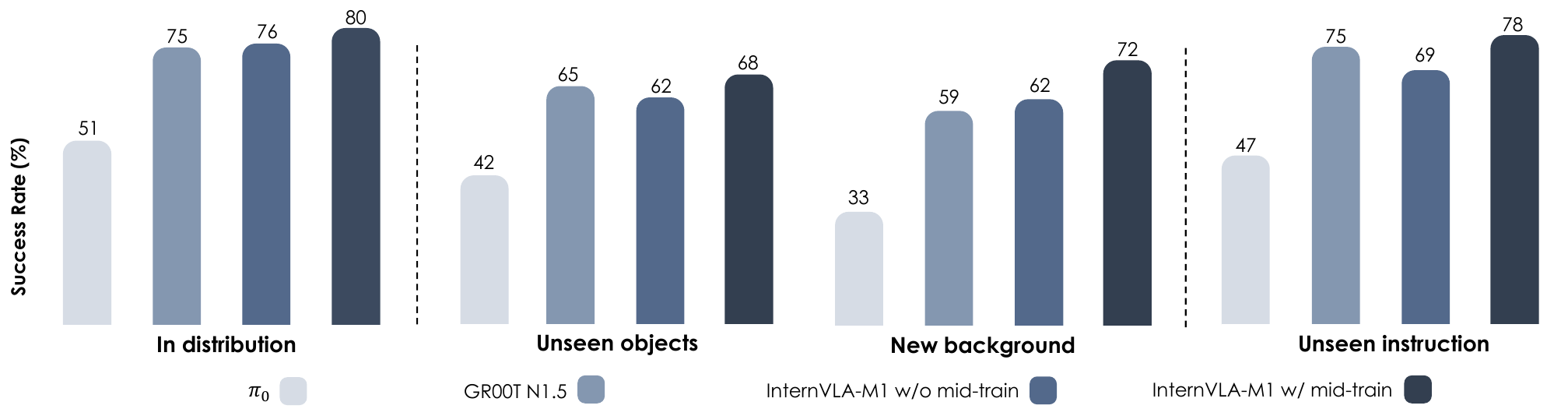}
    \caption{Result comparison of 200 simulated benchmarks in instruction-following pick-and-place.}
    \label{fig:sim_evaluation_200}
\end{figure}

\noindent
\textbf{Result analysis.}
As shown in~\Cref{fig:large_scale_simbench_200}, we evaluate {\frameworkname} under four generalization settings: In-distribution, Unseen Object, New Background, and Unseen Instruction. For each setting, we report two variants of the model: \emph{w/o mid-train}, which is fine-tuned using only five trajectories per task, and \emph{w/ mid-train}, which is additionally mid-trained on InternData M1 prior to fine-tuning. The results, summarized in \Cref{fig:sim_evaluation_200}, show that across all settings, both variants outperform the baseline $\pi_0$, while {\frameworkname~w/ mid-train} consistently surpasses GR00T N1.5. Although {\frameworkname~w/o mid-train} exhibits slight variance in certain settings, the mid-trained variant achieves a consistent advantage, with an average gain of +6.2\% over GR00T N1.5.

The performance on unseen objects highlights the benefit of simulation-enhanced visual generalization, enabling the model to handle novel instances beyond the training distribution. When evaluated under new backgrounds with randomized textures and layouts, both variants maintain strong performance, and the improvements from mid-training indicate increased robustness to scene-level shifts. Furthermore, under paraphrased instructions involving attribute-level or commonsense rewrites, {\frameworkname~w/ mid-train} demonstrates reliable instruction grounding, reflecting strong language generalization beyond templated expressions.

\subsubsection{Evaluation in Real-world Cluttered-scene Pick-and-Place}
\label{sec:real_results}

\begin{figure}[ht!]
    \centering
    \includegraphics[width=1.\textwidth]{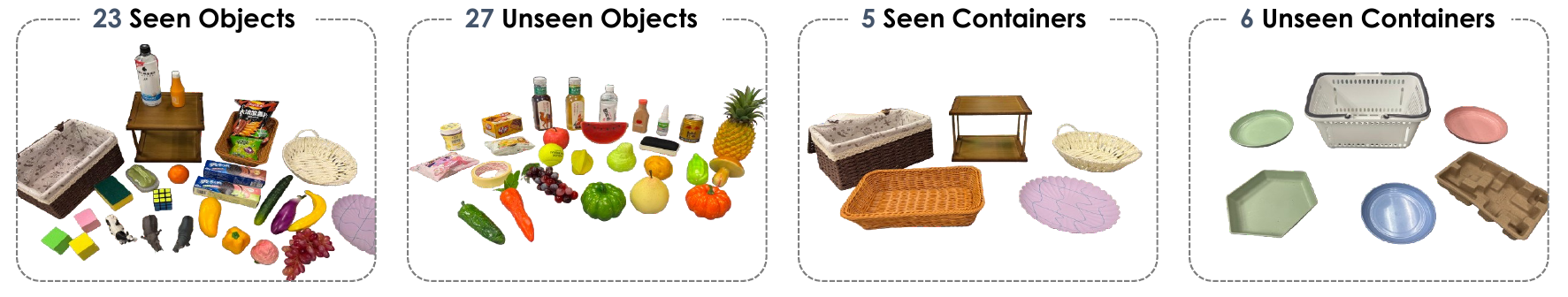}
    \caption{{Overview of objects and containers used in instruction-following pick-and-place.}}
    \label{fig:real-world-objects-containers}
\end{figure}

\noindent\textbf{Experimental setup.} 
To evaluate our model’s instruction-following capability in real-world scenarios, we employ a Franka Research 3 robotic arm equipped with a Robotiq 2F-85 gripper. The setup includes two Intel RealSense D435 cameras for RGB visual input—one mounted on the end-effector and another positioned at a rear, third-person perspective. We assess the model across a variety of manipulation tasks, including short-range pick-and-place, long-horizon object sorting, drawer opening/closing, and sandwich assembly.
For quantitative evaluation, we design a real-world object-sorting benchmark consisting of single-horizon pick-and-place tasks within a $60\times90$ cm tabletop workspace. The benchmark features 23 seen objects and 5 seen containers (listed in \Cref{fig:real-world-objects-containers}). In each episode, three containers are fixed at designated positions, while diverse objects are scattered randomly among them. The model must follow natural language instructions to pick specific objects and place them into the correct containers.
To support post-training, we collect 6 hours of teleoperated demonstrations using only objects and containers from the predefined “seen” set. We compare two variants of {\frameworkname}, \emph{w/o co-train} and \emph{w/ co-train}, against GR00T N1.5 and $\pi_0$ across five evaluation regimes on this benchmark. 
\emph{{\frameworkname} w/o co-train} is fine-tuned solely on real-world demonstrations, while \emph{{\frameworkname} w/ co-train} jointly trains on both the real-world data and the simulation dataset InternData-M1. 
The two RGB views are resized to $224 \times 224$ and used as model inputs. 
For both variants of our model, we fine-tune each on individual suites independently using 16 A100 GPUs, with a batch size of 256 and an action chunk size of 16. Training is performed for 20K steps. 
All models, including baselines, are trained and executed using delta end-effector space control in real-world experiments.

\begin{figure}[ht!]
    \centering
    \includegraphics[width=1.\textwidth]{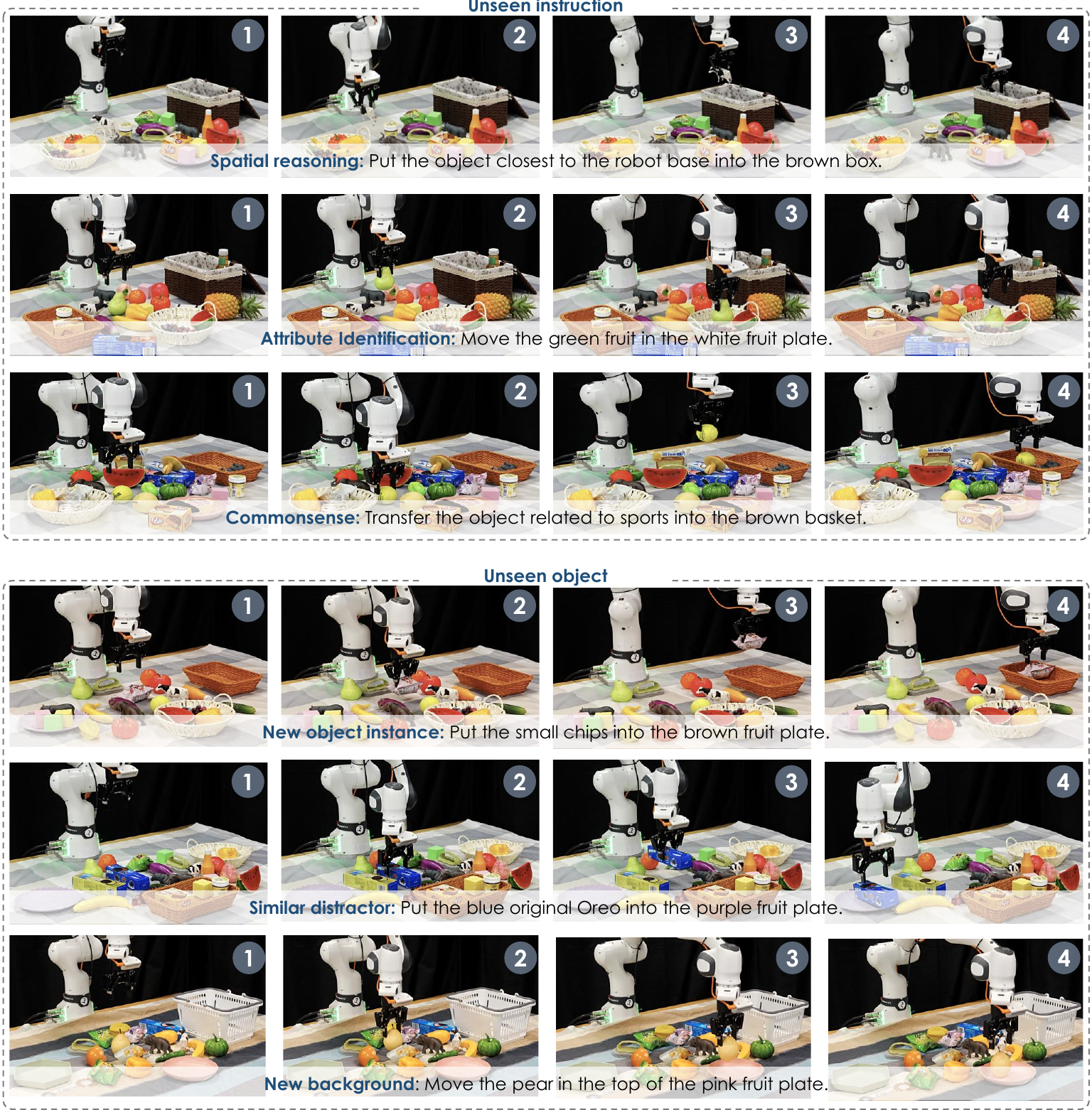}
    \caption{Evaluation settings showcase for real-world instruction-following manipulations.}
    \label{fig:real_world_evaluation_settings}
\end{figure}

\noindent\textbf{Evaluation settings.}
To evaluate generalization, we further partition all available object and container assets into disjoint seen and unseen sets, as illustrated in \Cref{fig:real-world-objects-containers}. Only the seen set is included in the training data, while both seen and unseen sets are evaluated during testing to measure the model’s ability to generalize to novel objects.
As shown in~\Cref{fig:real_world_evaluation_settings}, we evaluate instruction-following capabilities of various models on real-world pick-and-place tasks under the below settings: In-Distribution, Unseen Objects, Unseen Object Position, Unseen Object Orientation, and Unseen Instructions. We report success rate, defined as the fraction of trials in which the specified object is placed into the designated container. Higher SR indicates better performance.
For each model, we conducted a total of 300 rollout evaluations. Each trial corresponds to one or more testing settings, and we ensured that each individual setting was evaluated at least 50 times. To ensure fair comparisons across models, we fixed the positions of the objects and containers for each task during testing.

\begin{figure}[ht!]
    \centering
    \includegraphics[width=1.\textwidth]{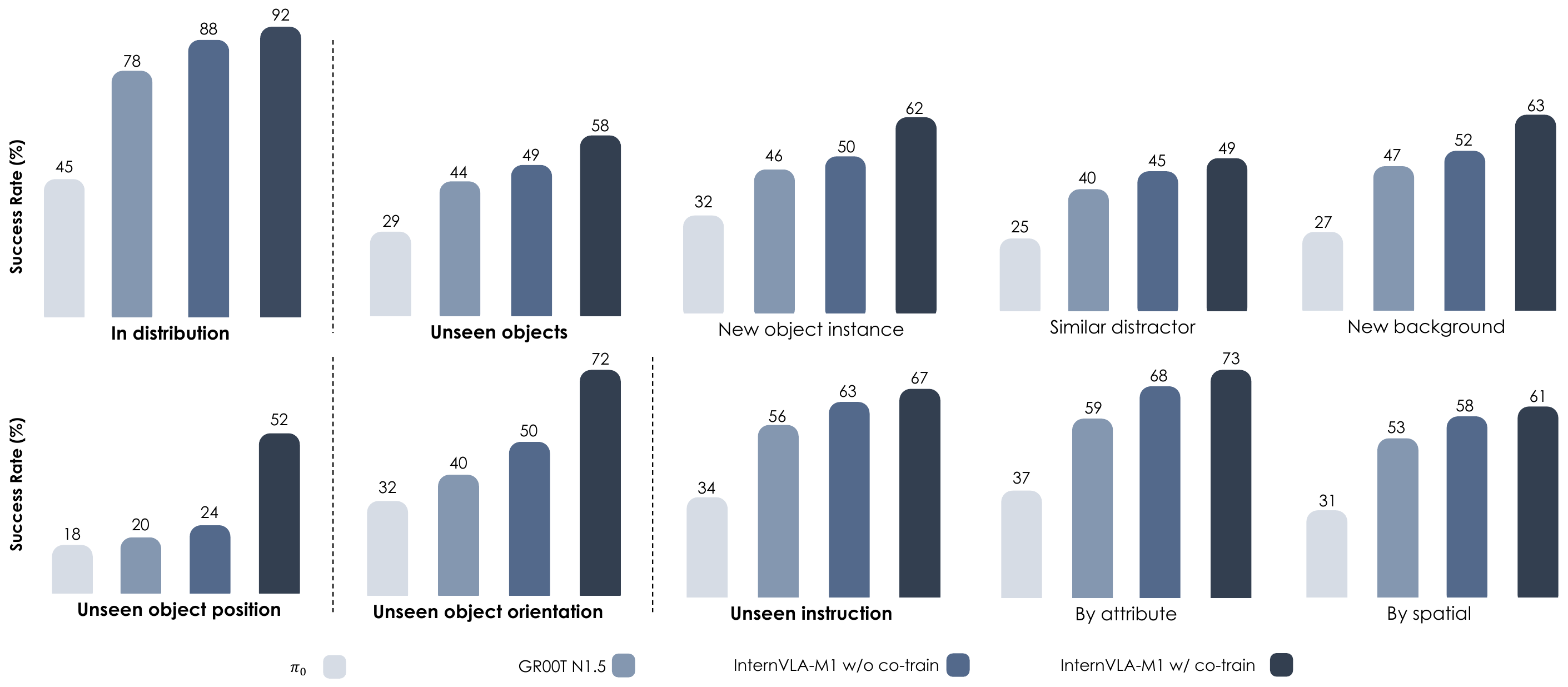}
    \caption{Result comparison in real-world instruction-following pick-and-place. }
    \label{fig:real world instruction following}
\end{figure}

\noindent\textbf{Result analysis.}
As shown in \Cref{fig:real world instruction following}, both variants of {\frameworkname} demonstrate superior performance under the in-distribution setting, consistently outperforming GR00T N1.5 and $\pi_0$ when evaluated on objects and containers seen during training. This indicates strong instruction-following capabilities within familiar contexts. Beyond this, the inclusion of Interndata-M1 during co-training significantly enhances the model’s visual generalization, enabling improved performance on novel objects not encountered during training. This suggests that synthetic data serves as an effective complement to limited real-world demonstrations. Additionally, because real-world data collection cannot exhaustively cover the spatial workspace, simulation data enriches the distribution of object positions and orientations. This leads to substantially better generalization to unseen configurations in terms of both object placement and pose. Finally, {\frameworkname} maintains robust performance when given novel instructions, highlighting its ability to generalize across diverse linguistic expressions beyond those seen during training.

\vspace{-1.3 em}
\subsubsection{Evaluation in Long-horizon and Reasoning Manipulation}
\label{sec:exp_long_task}

\begin{figure}[ht!]
    \centering
    \includegraphics[width=1.\textwidth]{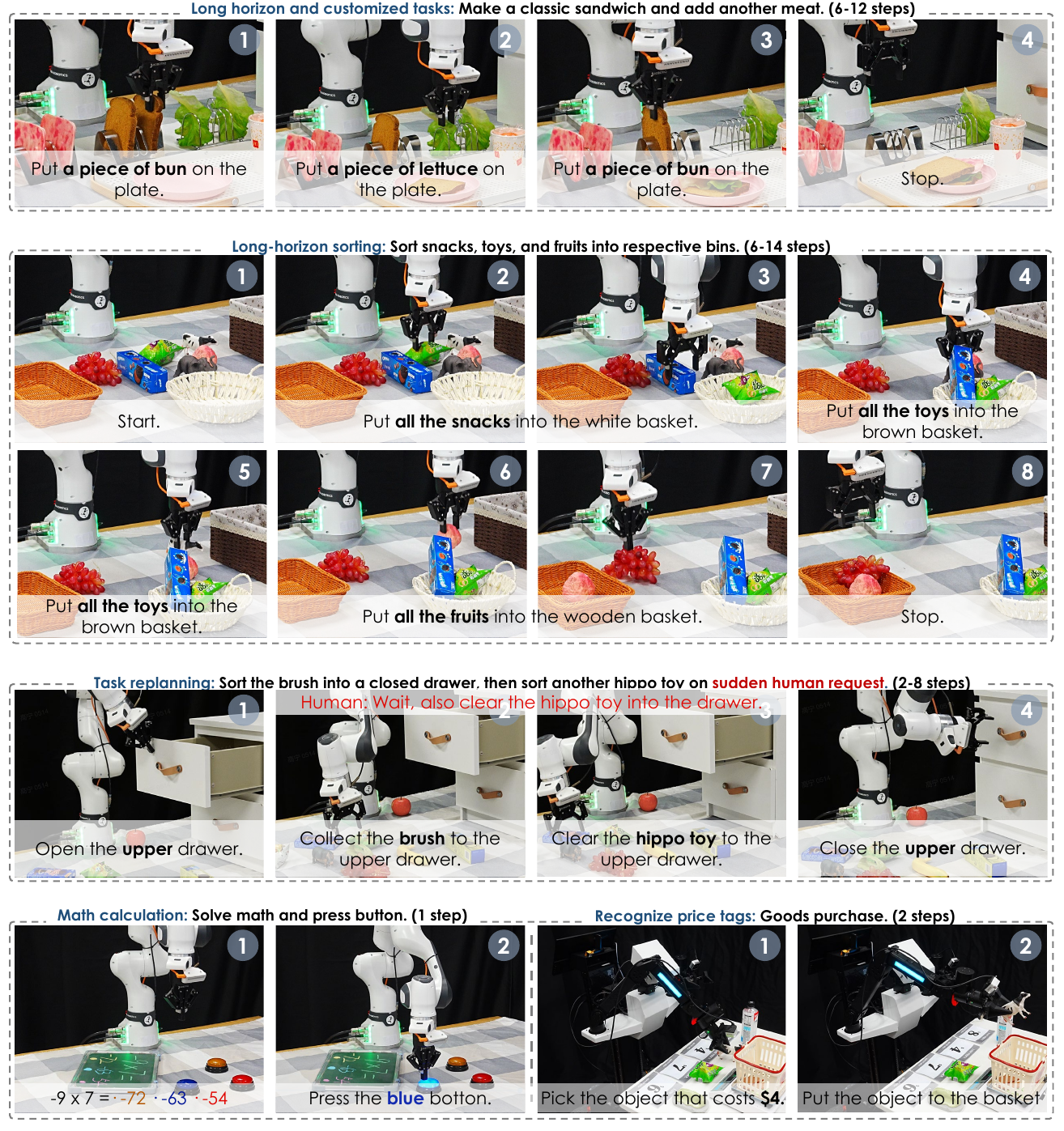}
    \caption{Showcase for long-horizon instruction-following manipulation.}
    \label{fig:real-world-long-horizion-showcase}
\end{figure}

A key strength of our dual-system framework is its ability to leverage a high-level planner (System 2) to decompose long-horizon, reasoning-heavy tasks into a sequence of atomic actions, which are then robustly executed by a low-level action model (System 1). To evaluate this capability, we design a series of tasks that require not only multi-step planning but also the ability to reason about object attributes, monitor progress, and adapt to changes. As illustrated in \Cref{fig:real-world-long-horizion-showcase}, these include:

\begin{itemize}[leftmargin=0.15in]

\item \textbf{Desktop Sorting.}
The Franka robot is tasked with sorting objects into containers based on high-level semantic categories, aiming to ensure that all items on the desktop are eventually placed into the correct containers. Both objects and containers are scattered within a 60×90 cm region in front of the robot base. The setup includes five seen containers and five object categories: \textit{fruits, toys, vegetables, bottles}, and \textit{snacks}. Each evaluation instance involves sorting objects from one to three categories into their respective containers. Each trial consists of three pick-and-place actions, and we report success rates consistent with the metric used for pick-and-place under clustered environments.

\item \textbf{Sorting Items into Drawers.}
The Franka robot is required to (i) open a designated drawer (either lower or upper), (ii) place the target objects into it, and (iii) close the drawer. This task demands precise temporal reasoning and articulated manipulation. The objects are placed within a 35×35 cm area located to the front-right of the robot base. We report stepwise execution success, where a step is considered valid only if all preceding steps have succeeded.

\item \textbf{Making Sandwiches.}
The Franka robot is instructed to assemble sandwiches following a predefined meal recipe. Ingredients and plates are placed within a 50×70 cm region in front of the robot base. We define five types of sandwich recipes as the seen set: 
[\,bread--lettuce--bread\,], [\,bread--lettuce--meat--bread\,], [\,bread--meat--lettuce--meat--bread\,], [\,bread--meat--meat--bread\,], and [\,bread--meat--bread\,]. 
We report success rates on both the seen set and an unseen set involving real-time environment interaction, using the same success definition as in the drawer sorting task.

\item \textbf{Math Calculation.}
The Franka robot is prompted to solve a math problem and press the color-coded button (red, yellow, or blue) that corresponds to the correct answer based on arithmetic reasoning. The buttons are randomly placed within a 40×40 cm area in front of the robot base. 

\item \textbf{Goods Purchase.}
The ARX LIFT2 dual-arm robot is tasked with identifying and placing into a basket the object bearing the correct price tag, given a numerical cue ranging from 1 to 9. We report the success rate of correctly placing the item corresponding to the queried price into the basket.

\end{itemize}

\begin{figure}[ht!]
    \centering
    \includegraphics[width=1.\textwidth]{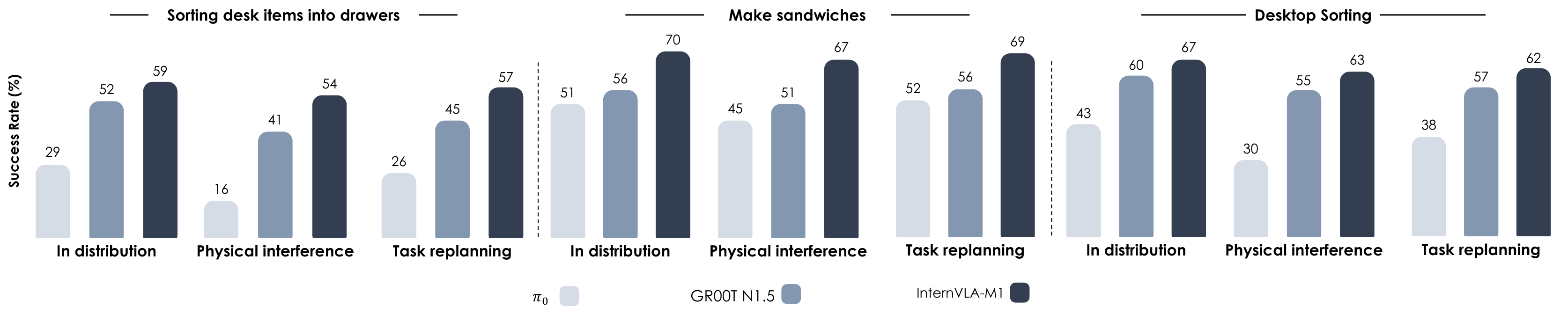}
    \caption{Result comparison in real-world long-horizon task planning for manipulation.}
    \label{fig:real_world_long horizon}
\end{figure}

\noindent
\textbf{Experimental setup.}
To support fine-grained training for these long-horizon tasks, we collect a total of 22 hours of high-quality long-horizon and reasoning teleoperated demonstrations, amounting to approximately 500 demonstrations per task. Each collected trajectory is segmented into \textit{subtasks} and annotated with corresponding atomic actions. For example, a “make a classic sandwich” task is decomposed into four subtasks:
(1) “Put a piece of bun on the plate.” →
(2) “Put a piece of meat on the plate.” →
(3) “Put a piece of lettuce on the plate.” →
(4) “Put a piece of bun on the plate.”
Each sub-instruction is paired with a specific segment of the demonstration.
To enable subtask-level transition, we introduce zero-action vectors padding after each subtask segment.
This allows the model to stop upon subtask completion and then be prompted to predict the transition to the next subtask. In addition, to improve temporal consistency and ensure smooth inference, we remove frames in which the robot arm exhibits clear pauses or idle behavior.
In contrast to prior VLA models that depend on an additional VLM to serve as a task planner for long-horizon or reasoning-intensive tasks, our unified model architecture is trained jointly on multimodal inputs encompassing task decomposition, subtask identification, numerical reasoning, and action supervision. This joint training paradigm enables a single model to seamlessly integrate task planning, reasoning, and action prediction in an end-to-end fashion.
The training recipe follows that of real-world short-horizon pick-and-place tasks. As shown in \Cref{tab:open_loop_task_comparison}, despite GPT-5’s strong reasoning capability, additional post-training of the unified model notably improves performance on long-horizon and reasoning-intensive tasks, underscoring the necessity of post-training for effective high-level task planning.

\begin{table}[ht!]
\small
\centering
\renewcommand{\arraystretch}{1.15}
\setlength{\tabcolsep}{6pt}
\begin{tabular}{l *{5}{c}}
\toprule
\multirow{3}{*}{\textbf{Models}} & \multicolumn{3}{c}{\textbf{Long-horizon tasks}} & \multicolumn{2}{c}{\textbf{Reasoning tasks}} \\
\cmidrule(lr){2-4} \cmidrule(lr){5-6}
 & \makecell[c]{Sort into\\Drawers} & \makecell[c]{Make\\Sandwiches} & \makecell[c]{Desktop\\Sorting} & \makecell[c]{Math\\calculation} & \makecell[c]{Goods\\ Purchase} \\
\midrule
\textbf{Gemini-2.5 Pro} & 57 & 62 & \underline{83} & 53 & 61 \\
\textbf{GPT-5} & \underline{75} & 67 & 62 & \underline{79} & \underline{82} \\
\textbf{GPT-4o} & 37 & 57 & 35 & 39 & 41 \\
\textbf{Qwen2.5-VL-72B} & 31 & \underline{71} & 34 & 33 & 29 \\
\textbf{Qwen2.5-VL-3B} & 30 & 49 & 52 & 41 & 38 \\ \midrule
\textbf{Ours-3B} & \textbf{90} & \textbf{91} & \textbf{91} & \textbf{93} & \textbf{92} \\
\bottomrule
\end{tabular}
\caption{Task scheduling performance of VLM planner in long-horizon and reasoning scenarios.}
\label{tab:open_loop_task_comparison}
\end{table}

\noindent
\textbf{Evaluation settings.}
We evaluate model performance under three distinct settings: In-distribution, Physical Interference, and Task Replanning:

\begin{itemize}[leftmargin=0.15in]
    \item \textbf{Physical interference.} External disturbances are introduced during task execution. For example, during the \textit{sorting items into drawers} task, the drawer is manually closed after the robot opens it, or the target object is displaced during grasping. This evaluates the model’s ability to perceive environmental changes and adapt accordingly.

    \item \textbf{Task replanning.} New instructions are issued mid-execution. For instance, after placing an object in the drawer but before closing it, the robot is told: “Also put the cow toy into the top drawer.” This tests the model’s ability to incorporate new subgoals and dynamically adjust its plan.
\end{itemize}

\noindent\textbf{Results analysis.}
As shown in~\Cref{fig:real_world_long horizon}, across long-horizon tasks, {\frameworkname} consistently outperforms the baselines, enabled by its unified subtask planning mechanism. In the in-distribution setting, it achieves more reliable execution than GR00T N1.5 and $\pi_0$, showing stronger grounding of high-level goals into actionable steps. Under physical interference, the model demonstrates robust adaptability: for example, in desktop sorting when containers are unexpectedly moved, {\frameworkname} can track the new container locations and complete the placement. Moreover, when task replanning is required, such as when additional instructions are introduced during execution, {\frameworkname} is able to revise its subtask sequence on the fly and continue with correct actions. This adaptability leads to minimal performance degradation under stress conditions, while the baselines exhibit much larger declines, underscoring the model’s resilience to dynamic environments and shifting instructions.

\vspace{-1.3 em}
\section{Related work}
\label{sec:related}

\noindent\textbf{Hierarchical robot system.} A key challenge in embodied AI is bridging high-level instructions with low-level actions, often addressed by generating intermediate representations (IRs) that range from formal symbolic structures to learned embeddings~\cite{xie2019embedding}. Inspired by Chain-of-Thought (CoT) reasoning, many approaches train vision-language-action (VLA) models to generate explicit textual plans before acting, which improves both interpretability and performance on complex tasks~\cite{ecot}. Beyond textual plans, research has explored more structured or physically grounded IRs. Historically, systems relied on direct perceptual outputs, such as bounding boxes from object detectors for manipulation~\cite{griffin2023mobile}, specific 3D points for grasp planning from point clouds~\cite{ten2017using}, or dense correspondence fields derived from self-supervised features learned for control, such as DINO features~\cite{laskin2020curl, nair2022r3m}. Some methods construct persistent 3D scene graphs as comprehensive world models that LLMs can query to ground long-horizon plans~\cite{rana2023sayplan}. Others emphasize action-centric IRs, for example by conditioning policies on visual affordances that specify the robot’s end-effector pose at key moments in a task~\cite{nasiriany2024rt}. A growing trend involves generating explicit spatial localizers directly consumable by robot controllers~\cite{huang2025roboground,rt-trajectory,li2025hamster}. Large-scale foundation models~\cite{team2025robobrain, vebrain} unify perception and planning by outputting not only plans but also affordance predictions as bounding boxes. For tasks requiring higher precision, specialized models such as RoboRefer~\cite{zhou2025roborefer} employ dedicated architectures and reinforcement learning to predict exact 3D coordinates from complex spatial language. In contrast, our method provides a unified latent modeling framework that integrates spatial guidance into downstream action training, enabling end-to-end optimization with direct feedback from real-world deployment. 

\noindent\textbf{Embodied reasoning and planning in VLA.}
Chain-of-Thought prompting has proven effective for improving reasoning in large language models~\cite{wei2022chain}, and its success has inspired extensions to embodied AI. In Vision-Language-Action (VLA) models, generating intermediate reasoning steps before acting enables agents to handle complex, long-horizon tasks. Early approaches emphasized linguistic reasoning: ECOT~\cite{ecot} elicits explicit text-based plans and sub-tasks to enhance performance and interpretability; RT-H~\cite{rth} introduces a fine-grained “action language” for hierarchical policies and human intervention; InstructVLA~\cite{instructvla} jointly optimizes reasoning and action through VLA-IT, improving generalization; OneTwoVLA~\cite{lin2025onetwovla} adaptively alternates between “thinking” and execution; RAD~\cite{clark2025action} leverages action-free human videos to derive reasoning guides; and $\pi_{0.5}$~\cite{pi_05} trains on heterogeneous data before fine-tuning for subtask prediction. Later work has also explored visual and spatial modalities, such as graph-based representations for spatial reasoning~\cite{huang2025graphcot}. Despite their differences, these approaches all generate intermediate steps such as textual, visual, or spatial representations during inference. While effective, this adds computational overhead. In contrast, we propose a post-training phase that directly unlocks the VLM’s intrinsic reasoning capacity, removing the need for explicit generative reasoning.

\noindent\textbf{Generalist robot policy.} Recent research in general-purpose robotics has seen the emergence of several mainstream technical paradigms. Monolithic VLA models utilize a single end-to-end network to directly map multimodal inputs to tokenized low-level actions, as demonstrated by systems~\cite{RT-2, openvla, magma, molmoact}. In contrast, unified architectures decouple high-level cognition from low-level action, allowing for greater modularity and interpretability. This category has seen extensive exploration~\cite{pi_0, cogact, li2025cronusvla} leveraging specialized generative models for action synthesis. Other notable approaches in this vein~\cite{pi_05, song2025hume, chatvla2, instructvla, smolvla, GR3}, which uses an LLM to break down high-level language commands into intermediate action plans. A third paradigm is based on world models, which learn a predictive model of the environment's dynamics to enable planning and control. These models allow for simulating future outcomes, often facilitating planning via search in a learned latent space or by conditioning a separate policy. While powerful, this approach can be computationally intensive. Representative works~\cite{ye2025lapa, bjorck2025gr00t, UVA, worldvla, liao2025genie, seer, bu2025univla, baaiunivla, f1vla} exemplify this forward-predictive approach to decision-making. Our model adopts a typical dual-system approach, building upon the VLA with unified architectures, then introducing additional planning design, thereby achieving better adaptability to real-world environments.

\vspace{-1.3 em}
\section{Discussion and conclusion}
\label{sec:discussion}

In this work, we presented InternVLA-M1, a unified vision-language-action framework that leverages spatial grounding priors to bridge high-level multimodal reasoning with low-level robotic execution. By combining large-scale multimodal pre-training with spatially guided post-training, our model effectively transfers spatially grounded understanding into embodied control, achieving strong generalization to unseen objects, instructions, and environments. Extensive evaluations across simulation and real-world settings demonstrate that InternVLA-M1 surpasses existing VLA models and specialized systems in instruction following, long-horizon manipulation, and multimodal grounding, highlighting spatial reasoning as a unifying substrate for scalable and reliable generalist robots.

\newpage
\bibliography{refs}

\begin{thebibliography}{92}
\providecommand{\natexlab}[1]{#1}
\providecommand{\url}[1]{\texttt{#1}}
\expandafter\ifx\csname urlstyle\endcsname\relax
  \providecommand{\doi}[1]{doi: #1}\else
  \providecommand{\doi}{doi: \begingroup \urlstyle{rm}\Url}\fi

\bibitem[AI(2024)]{helix}
F.~AI.
\newblock Helix, 2024.
\newblock URL \url{https://www.figure.ai/news/helix}.

\bibitem[Bai et~al.(2025{\natexlab{a}})Bai, Chen, Liu, Wang, Ge, Song, Dang, Wang, Wang, Tang, Zhong, Zhu, Yang, Li, Wan, Wang, Ding, Fu, Xu, Ye, Zhang, Xie, Cheng, Zhang, Yang, Xu, Lin, and ...]{qwen25vl}
S.~Bai, K.~Chen, X.~Liu, J.~Wang, W.~Ge, S.~Song, K.~Dang, P.~Wang, S.~Wang, J.~Tang, H.~Zhong, Y.~Zhu, M.~Yang, Z.~Li, J.~Wan, P.~Wang, W.~Ding, Z.~Fu, Y.~Xu, J.~Ye, X.~Zhang, T.~Xie, Z.~Cheng, H.~Zhang, Z.~Yang, H.~Xu, J.~Lin, and ...
\newblock Qwen2.5-vl technical report.
\newblock \emph{arXiv preprint arXiv:2502.13923}, 2025{\natexlab{a}}.

\bibitem[Bai et~al.(2025{\natexlab{b}})Bai, Chen, Liu, Wang, Ge, Song, Dang, Wang, Wang, Tang, et~al.]{bai2025qwen2}
S.~Bai, K.~Chen, X.~Liu, J.~Wang, W.~Ge, S.~Song, K.~Dang, P.~Wang, S.~Wang, J.~Tang, et~al.
\newblock Qwen2. 5-vl technical report.
\newblock \emph{arXiv preprint arXiv:2502.13923}, 2025{\natexlab{b}}.

\bibitem[Belkhale et~al.(2024)Belkhale, Ding, Xiao, Sermanet, Vuong, Tompson, Chebotar, Dwibedi, and Sadigh]{rth}
S.~Belkhale, T.~Ding, T.~Xiao, P.~Sermanet, Q.~Vuong, J.~Tompson, Y.~Chebotar, D.~Dwibedi, and D.~Sadigh.
\newblock Rt-h: Action hierarchies using language.
\newblock \emph{arXiv preprint arXiv:2403.01823}, 2024.

\bibitem[Bjorck et~al.(2025)Bjorck, Casta{\~n}eda, Cherniadev, Da, Ding, Fan, Fang, Fox, Hu, Huang, et~al.]{bjorck2025gr00t}
J.~Bjorck, F.~Casta{\~n}eda, N.~Cherniadev, X.~Da, R.~Ding, L.~Fan, Y.~Fang, D.~Fox, F.~Hu, S.~Huang, et~al.
\newblock Gr00t n1: An open foundation model for generalist humanoid robots.
\newblock \emph{arXiv preprint arXiv:2503.14734}, 2025.

\bibitem[Black et~al.(2024)Black, Brown, Driess, Esmail, Equi, Finn, Fusai, Groom, Hausman, Ichter, et~al.]{pi_0}
K.~Black, N.~Brown, D.~Driess, A.~Esmail, M.~Equi, C.~Finn, N.~Fusai, L.~Groom, K.~Hausman, B.~Ichter, et~al.
\newblock $\backslash pi_0 $: A vision-language-action flow model for general robot control.
\newblock \emph{arXiv preprint arXiv:2410.24164}, 2024.

\bibitem[Brohan et~al.(2022)Brohan, Brown, Carbajal, Chebotar, Dabis, Finn, Gopalakrishnan, Hausman, Herzog, Hsu, et~al.]{RT-1}
A.~Brohan, N.~Brown, J.~Carbajal, Y.~Chebotar, J.~Dabis, C.~Finn, K.~Gopalakrishnan, K.~Hausman, A.~Herzog, J.~Hsu, et~al.
\newblock Rt-1: Robotics transformer for real-world control at scale.
\newblock \emph{arXiv preprint arXiv:2212.06817}, 2022.

\bibitem[Brohan et~al.(2023)Brohan, Brown, Carbajal, Chebotar, Chen, Choromanski, Ding, Driess, Dubey, Finn, et~al.]{RT-2}
A.~Brohan, N.~Brown, J.~Carbajal, Y.~Chebotar, X.~Chen, K.~Choromanski, T.~Ding, D.~Driess, A.~Dubey, C.~Finn, et~al.
\newblock Rt-2: Vision-language-action models transfer web knowledge to robotic control.
\newblock \emph{arXiv preprint arXiv:2307.15818}, 2023.

\bibitem[Bu et~al.(2025{\natexlab{a}})Bu, Cai, Chen, Cui, Ding, Feng, Gao, He, Hu, Huang, et~al.]{bu2025agibot}
Q.~Bu, J.~Cai, L.~Chen, X.~Cui, Y.~Ding, S.~Feng, S.~Gao, X.~He, X.~Hu, X.~Huang, et~al.
\newblock Agibot world colosseo: A large-scale manipulation platform for scalable and intelligent embodied systems.
\newblock \emph{arXiv preprint arXiv:2503.06669}, 2025{\natexlab{a}}.

\bibitem[Bu et~al.(2025{\natexlab{b}})Bu, Yang, Cai, Gao, Ren, Yao, Luo, and Li]{bu2025univla}
Q.~Bu, Y.~Yang, J.~Cai, S.~Gao, G.~Ren, M.~Yao, P.~Luo, and H.~Li.
\newblock Univla: Learning to act anywhere with task-centric latent actions.
\newblock \emph{arXiv preprint arXiv:2505.06111}, 2025{\natexlab{b}}.

\bibitem[Cao et~al.(2025)Cao, Team, et~al.]{cao2025robobrain2}
M.~Cao, B.~R. Team, et~al.
\newblock Robobrain 2.0 technical report.
\newblock Technical report, Beijing Academy of Artificial Intelligence (BAAI), 2025.
\newblock arXiv preprint arXiv:2507.02029.

\bibitem[Cen et~al.(2025)Cen, Yu, Yuan, Jiang, Huang, Guo, Li, Song, Luo, Wang, et~al.]{worldvla}
J.~Cen, C.~Yu, H.~Yuan, Y.~Jiang, S.~Huang, J.~Guo, X.~Li, Y.~Song, H.~Luo, F.~Wang, et~al.
\newblock Worldvla: Towards autoregressive action world model.
\newblock \emph{arXiv preprint arXiv:2506.21539}, 2025.

\bibitem[Cheang et~al.(2025)Cheang, Chen, Cui, Hu, Huang, Kong, Li, Li, Liu, Ma, et~al.]{GR3}
C.~Cheang, S.~Chen, Z.~Cui, Y.~Hu, L.~Huang, T.~Kong, H.~Li, Y.~Li, Y.~Liu, X.~Ma, et~al.
\newblock Gr-3 technical report.
\newblock \emph{arXiv preprint arXiv:2507.15493}, 2025.

\bibitem[Chen et~al.(2015)Chen, Fang, Lin, and et~al.]{chen2015microsoft}
X.~Chen, H.~Fang, T.-Y. Lin, and et~al.
\newblock Microsoft coco captions: Data collection and evaluation server, 2015.

\bibitem[Chen et~al.(2024)Chen, Wu, Wang, Su, Chen, Xing, Zhong, Zhang, Zhu, Lu, et~al.]{chen2024internvl}
Z.~Chen, J.~Wu, W.~Wang, W.~Su, G.~Chen, S.~Xing, M.~Zhong, Q.~Zhang, X.~Zhu, L.~Lu, et~al.
\newblock Internvl: Scaling up vision foundation models and aligning for generic visual-linguistic tasks.
\newblock In \emph{Proceedings of the IEEE/CVF Conference on Computer Vision and Pattern Recognition}, pages 24185--24198, 2024.

\bibitem[Chi et~al.(2023)Chi, Feng, Du, Xu, Cousineau, Burchfiel, and Song]{chi2023diffusionpolicy}
C.~Chi, S.~Feng, Y.~Du, Z.~Xu, E.~Cousineau, B.~Burchfiel, and S.~Song.
\newblock Diffusion policy: Visuomotor policy learning via action diffusion.
\newblock In \emph{Proceedings of Robotics: Science and Systems (RSS)}, 2023.

\bibitem[Clark et~al.(2025)Clark, Mirchandani, Sadigh, and Belkhale]{clark2025action}
J.~Clark, S.~Mirchandani, D.~Sadigh, and S.~Belkhale.
\newblock Action-free reasoning for policy generalization.
\newblock \emph{arXiv preprint arXiv:2502.03729}, 2025.

\bibitem[Collaboration et~al.(2023)Collaboration, O'Neill, Rehman, Gupta, Maddukuri, Gupta, Padalkar, Lee, Pooley, Gupta, Mandlekar, Jain, Tung, Bewley, Herzog, Irpan, Khazatsky, Rai, Gupta, Wang, Kolobov, Singh, Garg, Kembhavi, Xie, Brohan, Raffin, Sharma, Yavary, Jain, Balakrishna, Wahid, Burgess-Limerick, Kim, Schölkopf, Wulfe, Ichter, Lu, Xu, Le, Finn, Wang, Xu, Chi, Huang, Chan, Agia, Pan, Fu, Devin, Xu, Morton, Driess, Chen, Pathak, Shah, Büchler, Jayaraman, Kalashnikov, Sadigh, Johns, Foster, Liu, Ceola, Xia, Zhao, Frujeri, Stulp, Zhou, Sukhatme, Salhotra, Yan, Feng, Schiavi, Berseth, Kahn, Yang, Wang, Su, Fang, Shi, Bao, Amor, Christensen, Furuta, Bharadhwaj, Walke, Fang, Ha, Mordatch, Radosavovic, Leal, Liang, Abou-Chakra, Kim, Drake, Peters, Schneider, Hsu, Vakil, Bohg, Bingham, Wu, Gao, Hu, Wu, Wu, Sun, Luo, Gu, Tan, Oh, Wu, Lu, Yang, Malik, Silvério, Hejna, Booher, Tompson, Yang, Salvador, Lim, Han, Wang, Rao, Pertsch, Hausman, Go, Gopalakrishnan, Goldberg, Byrne, Oslund, Kawaharazuka, Black,
  Lin, Zhang, Ehsani, Lekkala, Ellis, Rana, Srinivasan, Fang, Singh, Zeng, Hatch, Hsu, Itti, Chen, Pinto, Fei-Fei, Tan, Fan, Ott, Lee, Weihs, Chen, Lepert, Memmel, Tomizuka, Itkina, Castro, Spero, Du, Ahn, Yip, Zhang, Ding, Heo, Srirama, Sharma, Kim, Kanazawa, Hansen, Heess, Joshi, Suenderhauf, Liu, Palo, Shafiullah, Mees, Kroemer, Bastani, Sanketi, Miller, Yin, Wohlhart, Xu, Fagan, Mitrano, Sermanet, Abbeel, Sundaresan, Chen, Vuong, Rafailov, Tian, Doshi, Mart{'i}n-Mart{'i}n, Baijal, Scalise, Hendrix, Lin, Qian, Zhang, Mendonca, Shah, Hoque, Julian, Bustamante, Kirmani, Levine, Lin, Moore, Bahl, Dass, Sonawani, Tulsiani, Song, Xu, Haldar, Karamcheti, Adebola, Guist, Nasiriany, Schaal, Welker, Tian, Ramamoorthy, Dasari, Belkhale, Park, Nair, Mirchandani, Osa, Gupta, Harada, Matsushima, Xiao, Kollar, Yu, Ding, Davchev, Zhao, Armstrong, Darrell, Chung, Jain, Kumar, Vanhoucke, Zhan, Zhou, Burgard, Chen, Chen, Wang, Zhu, Geng, Liu, Liangwei, Li, Pang, Lu, Ma, Kim, Chebotar, Zhou, Zhu, Wu, Xu, Wang, Bisk, Dou,
  Cho, Lee, Cui, Cao, Wu, Tang, Zhu, Zhang, Jiang, Li, Li, Iwasawa, Matsuo, Ma, Xu, Cui, Zhang, Fu, and Lin]{open_x_embodiment}
O.~X.-E. Collaboration, A.~O'Neill, A.~Rehman, A.~Gupta, A.~Maddukuri, A.~Gupta, A.~Padalkar, A.~Lee, A.~Pooley, A.~Gupta, A.~Mandlekar, A.~Jain, A.~Tung, A.~Bewley, A.~Herzog, A.~Irpan, A.~Khazatsky, A.~Rai, A.~Gupta, A.~Wang, A.~Kolobov, A.~Singh, A.~Garg, A.~Kembhavi, A.~Xie, A.~Brohan, A.~Raffin, A.~Sharma, A.~Yavary, A.~Jain, A.~Balakrishna, A.~Wahid, B.~Burgess-Limerick, B.~Kim, B.~Schölkopf, B.~Wulfe, B.~Ichter, C.~Lu, C.~Xu, C.~Le, C.~Finn, C.~Wang, C.~Xu, C.~Chi, C.~Huang, C.~Chan, C.~Agia, C.~Pan, C.~Fu, C.~Devin, D.~Xu, D.~Morton, D.~Driess, D.~Chen, D.~Pathak, D.~Shah, D.~Büchler, D.~Jayaraman, D.~Kalashnikov, D.~Sadigh, E.~Johns, E.~Foster, F.~Liu, F.~Ceola, F.~Xia, F.~Zhao, F.~V. Frujeri, F.~Stulp, G.~Zhou, G.~S. Sukhatme, G.~Salhotra, G.~Yan, G.~Feng, G.~Schiavi, G.~Berseth, G.~Kahn, G.~Yang, G.~Wang, H.~Su, H.-S. Fang, H.~Shi, H.~Bao, H.~B. Amor, H.~I. Christensen, H.~Furuta, H.~Bharadhwaj, H.~Walke, H.~Fang, H.~Ha, I.~Mordatch, I.~Radosavovic, I.~Leal, J.~Liang, J.~Abou-Chakra, J.~Kim,
  J.~Drake, J.~Peters, J.~Schneider, J.~Hsu, J.~Vakil, J.~Bohg, J.~Bingham, J.~Wu, J.~Gao, J.~Hu, J.~Wu, J.~Wu, J.~Sun, J.~Luo, J.~Gu, J.~Tan, J.~Oh, J.~Wu, J.~Lu, J.~Yang, J.~Malik, J.~Silvério, J.~Hejna, J.~Booher, J.~Tompson, J.~Yang, J.~Salvador, J.~J. Lim, J.~Han, K.~Wang, K.~Rao, K.~Pertsch, K.~Hausman, K.~Go, K.~Gopalakrishnan, K.~Goldberg, K.~Byrne, K.~Oslund, K.~Kawaharazuka, K.~Black, K.~Lin, K.~Zhang, K.~Ehsani, K.~Lekkala, K.~Ellis, K.~Rana, K.~Srinivasan, K.~Fang, K.~P. Singh, K.-H. Zeng, K.~Hatch, K.~Hsu, L.~Itti, L.~Y. Chen, L.~Pinto, L.~Fei-Fei, L.~Tan, L.~J. Fan, L.~Ott, L.~Lee, L.~Weihs, M.~Chen, M.~Lepert, M.~Memmel, M.~Tomizuka, M.~Itkina, M.~G. Castro, M.~Spero, M.~Du, M.~Ahn, M.~C. Yip, M.~Zhang, M.~Ding, M.~Heo, M.~K. Srirama, M.~Sharma, M.~J. Kim, N.~Kanazawa, N.~Hansen, N.~Heess, N.~J. Joshi, N.~Suenderhauf, N.~Liu, N.~D. Palo, N.~M.~M. Shafiullah, O.~Mees, O.~Kroemer, O.~Bastani, P.~R. Sanketi, P.~T. Miller, P.~Yin, P.~Wohlhart, P.~Xu, P.~D. Fagan, P.~Mitrano, P.~Sermanet,
  P.~Abbeel, P.~Sundaresan, Q.~Chen, Q.~Vuong, R.~Rafailov, R.~Tian, R.~Doshi, R.~Mart{'i}n-Mart{'i}n, R.~Baijal, R.~Scalise, R.~Hendrix, R.~Lin, R.~Qian, R.~Zhang, R.~Mendonca, R.~Shah, R.~Hoque, R.~Julian, S.~Bustamante, S.~Kirmani, S.~Levine, S.~Lin, S.~Moore, S.~Bahl, S.~Dass, S.~Sonawani, S.~Tulsiani, S.~Song, S.~Xu, S.~Haldar, S.~Karamcheti, S.~Adebola, S.~Guist, S.~Nasiriany, S.~Schaal, S.~Welker, S.~Tian, S.~Ramamoorthy, S.~Dasari, S.~Belkhale, S.~Park, S.~Nair, S.~Mirchandani, T.~Osa, T.~Gupta, T.~Harada, T.~Matsushima, T.~Xiao, T.~Kollar, T.~Yu, T.~Ding, T.~Davchev, T.~Z. Zhao, T.~Armstrong, T.~Darrell, T.~Chung, V.~Jain, V.~Kumar, V.~Vanhoucke, W.~Zhan, W.~Zhou, W.~Burgard, X.~Chen, X.~Chen, X.~Wang, X.~Zhu, X.~Geng, X.~Liu, X.~Liangwei, X.~Li, Y.~Pang, Y.~Lu, Y.~J. Ma, Y.~Kim, Y.~Chebotar, Y.~Zhou, Y.~Zhu, Y.~Wu, Y.~Xu, Y.~Wang, Y.~Bisk, Y.~Dou, Y.~Cho, Y.~Lee, Y.~Cui, Y.~Cao, Y.-H. Wu, Y.~Tang, Y.~Zhu, Y.~Zhang, Y.~Jiang, Y.~Li, Y.~Li, Y.~Iwasawa, Y.~Matsuo, Z.~Ma, Z.~Xu, Z.~J. Cui, Z.~Zhang,
  Z.~Fu, and Z.~Lin.
\newblock Open {X-E}mbodiment: Robotic learning datasets and {RT-X} models.
\newblock \url{https://arxiv.org/abs/2310.08864}, 2023.

\bibitem[Deitke et~al.(2024)Deitke, Clark, Lee, Tripathi, Yang, Park, Salehi, Muennighoff, Lo, Soldaini, Lu, Anderson, Bransom, Ehsani, Ngo, Chen, Patel, Yatskar, Callison-Burch, Head, Hendrix, Bastani, VanderBilt, Lambert, Chou, Chheda, Sparks, Skjonsberg, Schmitz, Sarnat, Bischoff, Walsh, Newell, Wolters, Gupta, Zeng, Borchardt, Groeneveld, Dumas, Nam, Lebrecht, Wittlif, Schoenick, Michel, Krishna, Weihs, Smith, Hajishirzi, Girshick, Farhadi, and Kembhavi]{pixmo2024}
M.~Deitke, C.~Clark, S.~Lee, R.~Tripathi, Y.~Yang, J.~S. Park, M.~Salehi, N.~Muennighoff, K.~Lo, L.~Soldaini, J.~Lu, T.~Anderson, E.~Bransom, K.~Ehsani, H.~Ngo, Y.~Chen, A.~Patel, M.~Yatskar, C.~Callison-Burch, A.~Head, R.~Hendrix, F.~Bastani, E.~VanderBilt, N.~Lambert, Y.~Chou, A.~Chheda, J.~Sparks, S.~Skjonsberg, M.~Schmitz, A.~Sarnat, B.~Bischoff, P.~Walsh, C.~Newell, P.~Wolters, T.~Gupta, K.-H. Zeng, J.~Borchardt, D.~Groeneveld, J.~Dumas, C.~Nam, S.~Lebrecht, C.~Wittlif, C.~Schoenick, O.~Michel, R.~Krishna, L.~Weihs, N.~A. Smith, H.~Hajishirzi, R.~Girshick, A.~Farhadi, and A.~Kembhavi.
\newblock Molmo and pixmo: Open weights and open data for state-of-the-art multimodal models.
\newblock \emph{arXiv preprint arXiv:2409.17146}, 2024.

\bibitem[Driess et~al.(2025)Driess, Springenberg, Ichter, Yu, Li-Bell, Pertsch, Ren, Walke, Vuong, Shi, et~al.]{pi_05KI}
D.~Driess, J.~T. Springenberg, B.~Ichter, L.~Yu, A.~Li-Bell, K.~Pertsch, A.~Z. Ren, H.~Walke, Q.~Vuong, L.~X. Shi, et~al.
\newblock Knowledge insulating vision-language-action models: Train fast, run fast, generalize better.
\newblock \emph{arXiv preprint arXiv:2505.23705}, 2025.

\bibitem[Fang et~al.(2023)Fang, Wang, Fang, Gou, Liu, Yan, Liu, Xie, and Lu]{anygrasp}
H.-S. Fang, C.~Wang, H.~Fang, M.~Gou, J.~Liu, H.~Yan, W.~Liu, Y.~Xie, and C.~Lu.
\newblock Anygrasp: Robust and efficient grasp perception in spatial and temporal domains.
\newblock \emph{IEEE Transactions on Robotics}, 2023.

\bibitem[Gao et~al.(2025)Gao, Chen, Yang, Chen, Tian, Li, Huang, Wang, Wang, and Pang]{gao2025genmanip}
N.~Gao, Y.~Chen, S.~Yang, X.~Chen, Y.~Tian, H.~Li, H.~Huang, H.~Wang, T.~Wang, and J.~Pang.
\newblock Genmanip: Llm-driven simulation for generalizable instruction-following manipulation.
\newblock In \emph{CVPR}, 2025.

\bibitem[Griffin(2023)]{griffin2023mobile}
B.~Griffin.
\newblock Mobile robot manipulation using pure object detection.
\newblock In \emph{Proceedings of the IEEE/CVF Winter Conference on Applications of Computer Vision}, pages 561--571, 2023.

\bibitem[Gu et~al.(2023{\natexlab{a}})Gu, Kirmani, Wohlhart, Lu, Arenas, Rao, Yu, Fu, Gopalakrishnan, Xu, et~al.]{rt-trajectory}
J.~Gu, S.~Kirmani, P.~Wohlhart, Y.~Lu, M.~G. Arenas, K.~Rao, W.~Yu, C.~Fu, K.~Gopalakrishnan, Z.~Xu, et~al.
\newblock Rt-trajectory: Robotic task generalization via hindsight trajectory sketches.
\newblock \emph{arXiv preprint arXiv:2311.01977}, 2023{\natexlab{a}}.

\bibitem[Gu et~al.(2023{\natexlab{b}})Gu, Xiang, Li, Ling, Liu, Mu, Tang, Tao, Wei, Yao, et~al.]{gu2023maniskill2}
J.~Gu, F.~Xiang, X.~Li, Z.~Ling, X.~Liu, T.~Mu, Y.~Tang, S.~Tao, X.~Wei, Y.~Yao, et~al.
\newblock Maniskill2: A unified benchmark for generalizable manipulation skills.
\newblock \emph{arXiv preprint arXiv:2302.04659}, 2023{\natexlab{b}}.

\bibitem[Huang et~al.(2024{\natexlab{a}})Huang, Lin, Hu, Wang, and Gao]{copa}
H.~Huang, F.~Lin, Y.~Hu, S.~Wang, and Y.~Gao.
\newblock Copa: General robotic manipulation through spatial constraints of parts with foundation models.
\newblock \emph{arXiv preprint arXiv:2403.08248}, 2024{\natexlab{a}}.

\bibitem[Huang et~al.(2025{\natexlab{a}})Huang, Cen, Tan, Quan, Huang, and Zhang]{huang2025graphcot}
H.~Huang, M.~Cen, K.~Tan, X.~Quan, G.~Huang, and H.~Zhang.
\newblock Graphcot-vla: A 3d spatial-aware reasoning vision-language-action model for robotic manipulation with ambiguous instructions.
\newblock \emph{arXiv preprint arXiv:2508.07650}, 2025{\natexlab{a}}.

\bibitem[Huang et~al.(2025{\natexlab{b}})Huang, Chen, Chen, Li, Han, Wang, Wang, Pang, and Zhao]{huang2025roboground}
H.~Huang, X.~Chen, Y.~Chen, H.~Li, X.~Han, Z.~Wang, T.~Wang, J.~Pang, and Z.~Zhao.
\newblock Roboground: Robotic manipulation with grounded vision-language priors.
\newblock In \emph{Proceedings of the Computer Vision and Pattern Recognition Conference}, pages 22540--22550, 2025{\natexlab{b}}.

\bibitem[Huang et~al.(2023)Huang, Wang, Zhang, Li, Wu, and Fei-Fei]{voxposer}
W.~Huang, C.~Wang, R.~Zhang, Y.~Li, J.~Wu, and L.~Fei-Fei.
\newblock Voxposer: Composable 3d value maps for robotic manipulation with language models.
\newblock \emph{arXiv preprint arXiv:2307.05973}, 2023.

\bibitem[Huang et~al.(2024{\natexlab{b}})Huang, Wang, Li, Zhang, and Fei-Fei]{rekep}
W.~Huang, C.~Wang, Y.~Li, R.~Zhang, and L.~Fei-Fei.
\newblock Rekep: Spatio-temporal reasoning of relational keypoint constraints for robotic manipulation.
\newblock \emph{arXiv preprint arXiv:2409.01652}, 2024{\natexlab{b}}.

\bibitem[Intelligence et~al.(2025)Intelligence, Black, Brown, Darpinian, Dhabalia, Driess, Esmail, Equi, Finn, Fusai, et~al.]{pi_05}
P.~Intelligence, K.~Black, N.~Brown, J.~Darpinian, K.~Dhabalia, D.~Driess, A.~Esmail, M.~Equi, C.~Finn, N.~Fusai, et~al.
\newblock $pi_{0.5}$: a vision-language-action model with open-world generalization.
\newblock \emph{arXiv preprint arXiv:2504.16054}, 2025.

\bibitem[Khazatsky et~al.(2024)Khazatsky, Pertsch, Nair, Balakrishna, Dasari, Karamcheti, Nasiriany, Srirama, Chen, Ellis, et~al.]{khazatsky2024droid}
A.~Khazatsky, K.~Pertsch, S.~Nair, A.~Balakrishna, S.~Dasari, S.~Karamcheti, S.~Nasiriany, M.~K. Srirama, L.~Y. Chen, K.~Ellis, et~al.
\newblock Droid: A large-scale in-the-wild robot manipulation dataset.
\newblock \emph{arXiv preprint arXiv:2403.12945}, 2024.

\bibitem[Kim et~al.(2024)Kim, Pertsch, Karamcheti, Xiao, Balakrishna, Nair, Rafailov, Foster, Lam, Sanketi, et~al.]{openvla}
M.~J. Kim, K.~Pertsch, S.~Karamcheti, T.~Xiao, A.~Balakrishna, S.~Nair, R.~Rafailov, E.~Foster, G.~Lam, P.~Sanketi, et~al.
\newblock Openvla: An open-source vision-language-action model.
\newblock \emph{arXiv preprint arXiv:2406.09246}, 2024.

\bibitem[Kim et~al.(2025)Kim, Finn, and Liang]{openvla-oft}
M.~J. Kim, C.~Finn, and P.~Liang.
\newblock Fine-tuning vision-language-action models: Optimizing speed and success.
\newblock \emph{arXiv preprint arXiv:2502.19645}, 2025.

\bibitem[Kirillov et~al.(2023)Kirillov, Mintun, Ravi, Mao, Rolland, Gustafson, Xiao, Whitehead, Berg, Lo, Doll{\'a}r, and Girshick]{sam}
A.~Kirillov, E.~Mintun, N.~Ravi, H.~Mao, C.~Rolland, L.~Gustafson, T.~Xiao, S.~Whitehead, A.~C. Berg, W.-Y. Lo, P.~Doll{\'a}r, and R.~Girshick.
\newblock Segment anything.
\newblock \emph{arXiv:2304.02643}, 2023.

\bibitem[Laskin et~al.(2020)Laskin, Srinivas, and Abbeel]{laskin2020curl}
M.~Laskin, A.~Srinivas, and P.~Abbeel.
\newblock Curl: Contrastive unsupervised representations for reinforcement learning.
\newblock In \emph{International conference on machine learning}, pages 5639--5650. PMLR, 2020.

\bibitem[Lee et~al.(2025)Lee, Duan, Fang, Deng, Liu, Li, Fang, Zhang, Wang, Lee, et~al.]{molmoact}
J.~Lee, J.~Duan, H.~Fang, Y.~Deng, S.~Liu, B.~Li, B.~Fang, J.~Zhang, Y.~R. Wang, S.~Lee, et~al.
\newblock Molmoact: Action reasoning models that can reason in space.
\newblock \emph{arXiv preprint arXiv:2508.07917}, 2025.

\bibitem[Li et~al.(2024{\natexlab{a}})Li, Zhang, Guo, Zhang, Li, Zhang, Zhang, Zhang, Li, Liu, et~al.]{li2024llava}
B.~Li, Y.~Zhang, D.~Guo, R.~Zhang, F.~Li, H.~Zhang, K.~Zhang, P.~Zhang, Y.~Li, Z.~Liu, et~al.
\newblock Llava-onevision: Easy visual task transfer.
\newblock \emph{arXiv preprint arXiv:2408.03326}, 2024{\natexlab{a}}.

\bibitem[Li et~al.(2024{\natexlab{b}})Li, Zhang, Guo, Zhang, Li, Zhang, Zhang, Zhang, Li, Liu, et~al.]{llavaov}
B.~Li, Y.~Zhang, D.~Guo, R.~Zhang, F.~Li, H.~Zhang, K.~Zhang, P.~Zhang, Y.~Li, Z.~Liu, et~al.
\newblock Llava-onevision: Easy visual task transfer.
\newblock \emph{arXiv preprint arXiv:2408.03326}, 2024{\natexlab{b}}.

\bibitem[Li et~al.(2025{\natexlab{a}})Li, Yang, Chen, Tian, Yang, Chen, Wang, Wang, Zhao, Lin, et~al.]{li2025cronusvla}
H.~Li, S.~Yang, Y.~Chen, Y.~Tian, X.~Yang, X.~Chen, H.~Wang, T.~Wang, F.~Zhao, D.~Lin, et~al.
\newblock Cronusvla: Transferring latent motion across time for multi-frame prediction in manipulation.
\newblock \emph{arXiv preprint arXiv:2506.19816}, 2025{\natexlab{a}}.

\bibitem[Li et~al.(2024{\natexlab{c}})Li, Liang, Wang, Luo, Chen, Liao, Wei, Deng, Xu, Zhang, et~al.]{cogact}
Q.~Li, Y.~Liang, Z.~Wang, L.~Luo, X.~Chen, M.~Liao, F.~Wei, Y.~Deng, S.~Xu, Y.~Zhang, et~al.
\newblock Cogact: A foundational vision-language-action model for synergizing cognition and action in robotic manipulation.
\newblock \emph{arXiv preprint arXiv:2411.19650}, 2024{\natexlab{c}}.

\bibitem[Li et~al.(2025{\natexlab{b}})Li, Gao, Sadigh, and Song]{UVA}
S.~Li, Y.~Gao, D.~Sadigh, and S.~Song.
\newblock Unified video action model.
\newblock \emph{arXiv preprint arXiv:2503.00200}, 2025{\natexlab{b}}.

\bibitem[Li et~al.(2023)Li, Du, Zhou, Wang, Zhao, and Wen]{li2023evaluating}
Y.~Li, Y.~Du, K.~Zhou, J.~Wang, W.~X. Zhao, and J.-R. Wen.
\newblock Evaluating object hallucination in large vision-language models.
\newblock \emph{arXiv preprint arXiv:2305.10355}, 2023.

\bibitem[Li et~al.(2025{\natexlab{c}})Li, Deng, Zhang, Jang, Memmel, Yu, Garrett, Ramos, Fox, Li, et~al.]{li2025hamster}
Y.~Li, Y.~Deng, J.~Zhang, J.~Jang, M.~Memmel, R.~Yu, C.~R. Garrett, F.~Ramos, D.~Fox, A.~Li, et~al.
\newblock Hamster: Hierarchical action models for open-world robot manipulation.
\newblock \emph{arXiv preprint arXiv:2502.05485}, 2025{\natexlab{c}}.

\bibitem[Liang et~al.(2019)Liang, Ma, Li, G{\"o}rner, Tang, Fang, Sun, and Zhang]{liang2019pointnetgpd}
H.~Liang, X.~Ma, S.~Li, M.~G{\"o}rner, S.~Tang, B.~Fang, F.~Sun, and J.~Zhang.
\newblock Pointnetgpd: Detecting grasp configurations from point sets.
\newblock In \emph{2019 International Conference on Robotics and Automation (ICRA)}, pages 3629--3635. IEEE, 2019.

\bibitem[Liao et~al.(2025)Liao, Zhou, Huang, Yang, Chen, Jiang, Hu, Cai, Liu, Luo, et~al.]{liao2025genie}
Y.~Liao, P.~Zhou, S.~Huang, D.~Yang, S.~Chen, Y.~Jiang, Y.~Hu, J.~Cai, S.~Liu, J.~Luo, et~al.
\newblock Genie envisioner: A unified world foundation platform for robotic manipulation.
\newblock \emph{arXiv preprint arXiv:2508.05635}, 2025.

\bibitem[Lin et~al.(2025)Lin, Nai, Hu, You, Zhao, and Gao]{lin2025onetwovla}
F.~Lin, R.~Nai, Y.~Hu, J.~You, J.~Zhao, and Y.~Gao.
\newblock Onetwovla: A unified vision-language-action model with adaptive reasoning.
\newblock \emph{arXiv preprint arXiv:2505.11917}, 2025.

\bibitem[Liu et~al.(2024)Liu, Fang, Abbeel, and Levine]{moka}
F.~Liu, K.~Fang, P.~Abbeel, and S.~Levine.
\newblock Moka: Open-vocabulary robotic manipulation through mark-based visual prompting.
\newblock In \emph{First Workshop on Vision-Language Models for Navigation and Manipulation at ICRA 2024}, 2024.

\bibitem[Lu et~al.(2023{\natexlab{a}})Lu, Fan, Deng, Liu, Li, and Wang]{lu2023vl}
Y.~Lu, Y.~Fan, B.~Deng, F.~Liu, Y.~Li, and S.~Wang.
\newblock Vl-grasp: a 6-dof interactive grasp policy for language-oriented objects in cluttered indoor scenes.
\newblock In \emph{2023 IEEE/RSJ International Conference on Intelligent Robots and Systems (IROS)}, pages 976--983. IEEE, 2023{\natexlab{a}}.

\bibitem[Lu et~al.(2023{\natexlab{b}})Lu, Fan, Deng, Liu, Li, and Wang]{vlgrasp}
Y.~Lu, Y.~Fan, B.~Deng, F.~Liu, Y.~Li, and S.~Wang.
\newblock Vl-grasp: a 6-dof interactive grasp policy for language-oriented objects in cluttered indoor scenes.
\newblock In \emph{2023 IEEE/RSJ International Conference on Intelligent Robots and Systems (IROS)}, pages 976--983. IEEE, 2023{\natexlab{b}}.

\bibitem[Luo et~al.(2025)Luo, Yang, Gong, Chen, Duan, Cui, Tong, Hou, Zhang, Chen, et~al.]{vebrain}
G.~Luo, G.~Yang, Z.~Gong, G.~Chen, H.~Duan, E.~Cui, R.~Tong, Z.~Hou, T.~Zhang, Z.~Chen, et~al.
\newblock Visual embodied brain: Let multimodal large language models see, think, and control in spaces.
\newblock \emph{arXiv preprint arXiv:2506.00123}, 2025.

\bibitem[Lv et~al.(2025)Lv, Kong, Li, Zeng, Qiu, Qu, Song, Chen, Deng, Wang, Nie, and Pang]{f1vla}
Q.~Lv, W.~Kong, H.~Li, J.~Zeng, Z.~Qiu, D.~Qu, H.~Song, Q.~Chen, X.~Deng, M.~Y. Wang, L.~Nie, and J.~Pang.
\newblock F1: A vision-language-action model bridging understanding and generation to actions.
\newblock 2025.
\newblock URL \url{https://arxiv.org/abs/2509.06951}.

\bibitem[Makoviychuk et~al.(2021)Makoviychuk, Wawrzyniak, Guo, Lu, Storey, Macklin, Hoeller, Rudin, Allshire, Handa, et~al.]{isaac}
V.~Makoviychuk, L.~Wawrzyniak, Y.~Guo, M.~Lu, K.~Storey, M.~Macklin, D.~Hoeller, N.~Rudin, A.~Allshire, A.~Handa, et~al.
\newblock Isaac gym: High performance gpu-based physics simulation for robot learning.
\newblock \emph{arXiv preprint arXiv:2108.10470}, 2021.

\bibitem[Mao et~al.(2016)Mao, Huang, Toshev, Camburu, Yuille, and Murphy]{mao2016generation}
J.~Mao, J.~Huang, A.~Toshev, O.~Camburu, A.~L. Yuille, and K.~Murphy.
\newblock Generation and comprehension of unambiguous object descriptions.
\newblock In \emph{IEEE Conference on Computer Vision and Pattern Recognition (CVPR)}, pages 11--20, 2016.

\bibitem[Nair et~al.(2022)Nair, Rajeswaran, Kumar, Finn, and Gupta]{nair2022r3m}
S.~Nair, A.~Rajeswaran, V.~Kumar, C.~Finn, and A.~Gupta.
\newblock R3m: A universal visual representation for robot manipulation.
\newblock \emph{arXiv preprint arXiv:2203.12601}, 2022.

\bibitem[Nasiriany et~al.(2024)Nasiriany, Kirmani, Ding, Smith, Zhu, Driess, Sadigh, and Xiao]{nasiriany2024rt}
S.~Nasiriany, S.~Kirmani, T.~Ding, L.~Smith, Y.~Zhu, D.~Driess, D.~Sadigh, and T.~Xiao.
\newblock Rt-affordance: Affordances are versatile intermediate representations for robot manipulation.
\newblock \emph{arXiv preprint arXiv:2411.02704}, 2024.

\bibitem[{Octo Model Team} et~al.(2024){Octo Model Team}, Ghosh, Walke, Pertsch, Black, Mees, Dasari, Hejna, Xu, Luo, Kreiman, Tan, Sanketi, Vuong, Xiao, Sadigh, Finn, and Levine]{octo}
{Octo Model Team}, D.~Ghosh, H.~Walke, K.~Pertsch, K.~Black, O.~Mees, S.~Dasari, J.~Hejna, C.~Xu, J.~Luo, T.~Kreiman, Y.~Tan, P.~Sanketi, Q.~Vuong, T.~Xiao, D.~Sadigh, C.~Finn, and S.~Levine.
\newblock Octo: An open-source generalist robot policy.
\newblock In \emph{Proceedings of Robotics: Science and Systems}, Delft, Netherlands, 2024.

\bibitem[Oquab et~al.(2023)Oquab, Darcet, Moutakanni, Vo, Szafraniec, Khalidov, Fernandez, Haziza, Massa, El-Nouby, et~al.]{oquab2023dinov2}
M.~Oquab, T.~Darcet, T.~Moutakanni, H.~Vo, M.~Szafraniec, V.~Khalidov, P.~Fernandez, D.~Haziza, F.~Massa, A.~El-Nouby, et~al.
\newblock Dinov2: Learning robust visual features without supervision.
\newblock \emph{arXiv preprint arXiv:2304.07193}, 2023.

\bibitem[Pertsch et~al.(2025)Pertsch, Stachowicz, Ichter, Driess, Nair, Vuong, Mees, Finn, and Levine]{pertsch2025fast}
K.~Pertsch, K.~Stachowicz, B.~Ichter, D.~Driess, S.~Nair, Q.~Vuong, O.~Mees, C.~Finn, and S.~Levine.
\newblock Fast: Efficient action tokenization for vision-language-action models.
\newblock \emph{arXiv preprint arXiv:2501.09747}, 2025.

\bibitem[Qi et~al.(2025)Qi, Zhang, Ding, Dong, Yu, Li, Xu, Li, He, Fan, Zhang, He, Gu, Jin, Ma, Zhang, Wang, and Yi]{qi2025sofar}
Z.~Qi, W.~Zhang, Y.~Ding, R.~Dong, X.~Yu, J.~Li, L.~Xu, B.~Li, X.~He, G.~Fan, J.~Zhang, J.~He, J.~Gu, X.~Jin, K.~Ma, Z.~Zhang, H.~Wang, and L.~Yi.
\newblock Sofar: Language-grounded orientation bridges spatial reasoning and object manipulation.
\newblock \emph{CoRR}, abs/2502.13143, 2025.
\newblock \doi{10.48550/ARXIV.2502.13143}.
\newblock URL \url{https://doi.org/10.48550/arXiv.2502.13143}.

\bibitem[Qu et~al.(2025)Qu, Song, Chen, Yao, Ye, Ding, Wang, Gu, Zhao, Wang, et~al.]{spatialvla}
D.~Qu, H.~Song, Q.~Chen, Y.~Yao, X.~Ye, Y.~Ding, Z.~Wang, J.~Gu, B.~Zhao, D.~Wang, et~al.
\newblock Spatialvla: Exploring spatial representations for visual-language-action model.
\newblock \emph{arXiv preprint arXiv:2501.15830}, 2025.

\bibitem[Radford et~al.(2021)Radford, Kim, Hallacy, Ramesh, Goh, Agarwal, Sastry, Askell, Mishkin, Clark, et~al.]{clip}
A.~Radford, J.~W. Kim, C.~Hallacy, A.~Ramesh, G.~Goh, S.~Agarwal, G.~Sastry, A.~Askell, P.~Mishkin, J.~Clark, et~al.
\newblock Learning transferable visual models from natural language supervision.
\newblock In \emph{International conference on machine learning}, pages 8748--8763. PMLR, 2021.

\bibitem[Raghu et~al.(2017)Raghu, Gilmer, Yosinski, and Sohl-Dickstein]{raghu2017svcca}
M.~Raghu, J.~Gilmer, J.~Yosinski, and J.~Sohl-Dickstein.
\newblock Svcca: Singular vector canonical correlation analysis for deep learning dynamics and interpretability.
\newblock \emph{Advances in neural information processing systems}, 30, 2017.

\bibitem[Rana et~al.(2023)Rana, Haviland, Garg, Abou-Chakra, Reid, and Suenderhauf]{rana2023sayplan}
K.~Rana, J.~Haviland, S.~Garg, J.~Abou-Chakra, I.~Reid, and N.~Suenderhauf.
\newblock Sayplan: Grounding large language models using 3d scene graphs for scalable robot task planning.
\newblock \emph{arXiv preprint arXiv:2307.06135}, 2023.

\bibitem[Shi et~al.(2025)Shi, Ichter, Equi, Ke, Pertsch, Vuong, Tanner, Walling, Wang, Fusai, et~al.]{hirobot}
L.~X. Shi, B.~Ichter, M.~Equi, L.~Ke, K.~Pertsch, Q.~Vuong, J.~Tanner, A.~Walling, H.~Wang, N.~Fusai, et~al.
\newblock Hi robot: Open-ended instruction following with hierarchical vision-language-action models.
\newblock \emph{arXiv preprint arXiv:2502.19417}, 2025.

\bibitem[Shukor et~al.(2025)Shukor, Aubakirova, Capuano, Kooijmans, Palma, Zouitine, Aractingi, Pascal, Russi, Marafioti, et~al.]{smolvla}
M.~Shukor, D.~Aubakirova, F.~Capuano, P.~Kooijmans, S.~Palma, A.~Zouitine, M.~Aractingi, C.~Pascal, M.~Russi, A.~Marafioti, et~al.
\newblock Smolvla: A vision-language-action model for affordable and efficient robotics.
\newblock \emph{arXiv preprint arXiv:2506.01844}, 2025.

\bibitem[Singh et~al.(2019)Singh, Natarjan, Shah, Jiang, Chen, Parikh, and Rohrbach]{singh2019towards}
A.~Singh, V.~Natarjan, M.~Shah, Y.~Jiang, X.~Chen, D.~Parikh, and M.~Rohrbach.
\newblock Towards vqa models that can read.
\newblock In \emph{Proceedings of the IEEE Conference on Computer Vision and Pattern Recognition}, pages 8317--8326, 2019.

\bibitem[Singh et~al.(2024)Singh, Yadav, Jain, Shi, Johnson, and Desai]{singh2024benchmarking}
S.~Singh, A.~Yadav, J.~Jain, H.~Shi, J.~Johnson, and K.~Desai.
\newblock Benchmarking object detectors with coco: A new path forward.
\newblock In \emph{European Conference on Computer Vision}, pages 279--295. Springer, 2024.

\bibitem[Song et~al.(2025)Song, Qu, Yao, Chen, Lv, Tang, Shi, Ren, Yao, Zhao, et~al.]{song2025hume}
H.~Song, D.~Qu, Y.~Yao, Q.~Chen, Q.~Lv, Y.~Tang, M.~Shi, G.~Ren, M.~Yao, B.~Zhao, et~al.
\newblock Hume: Introducing system-2 thinking in visual-language-action model.
\newblock \emph{arXiv preprint arXiv:2505.21432}, 2025.

\bibitem[Team et~al.(2025)Team, Cao, Tan, Ji, Lin, Li, Cao, Wang, Zhou, Han, et~al.]{team2025robobrain}
B.~R. Team, M.~Cao, H.~Tan, Y.~Ji, M.~Lin, Z.~Li, Z.~Cao, P.~Wang, E.~Zhou, Y.~Han, et~al.
\newblock Robobrain 2.0 technical report.
\newblock \emph{arXiv preprint arXiv:2507.02029}, 2025.

\bibitem[Ten~Pas and Platt(2017)]{ten2017using}
A.~Ten~Pas and R.~Platt.
\newblock Using geometry to detect grasp poses in 3d point clouds.
\newblock In \emph{Robotics Research: Volume 1}, pages 307--324. Springer, 2017.

\bibitem[Tian et~al.(2024)Tian, Yang, Zeng, Wang, Lin, Dong, and Pang]{seer}
Y.~Tian, S.~Yang, J.~Zeng, P.~Wang, D.~Lin, H.~Dong, and J.~Pang.
\newblock Predictive inverse dynamics models are scalable learners for robotic manipulation.
\newblock \emph{arXiv preprint arXiv:2412.15109}, 2024.

\bibitem[Wang et~al.(2024)Wang, Ren, Luo, Li, Yan, Chen, Wang, Li, Lu, Zhu, et~al.]{wang2024all}
W.~Wang, Y.~Ren, H.~Luo, T.~Li, C.~Yan, Z.~Chen, W.~Wang, Q.~Li, L.~Lu, X.~Zhu, et~al.
\newblock The all-seeing project v2: Towards general relation comprehension of the open world.
\newblock In \emph{European Conference on Computer Vision}, pages 471--490. Springer, 2024.

\bibitem[Wang et~al.(2025)Wang, Li, Wang, Zhang, Li, Chen, Wang, and Zhang]{baaiunivla}
Y.~Wang, X.~Li, W.~Wang, J.~Zhang, Y.~Li, Y.~Chen, X.~Wang, and Z.~Zhang.
\newblock Unified vision-language-action model.
\newblock \emph{arXiv preprint arXiv:2506.19850}, 2025.

\bibitem[Wei et~al.(2022)Wei, Wang, Schuurmans, Bosma, Xia, Chi, Le, Zhou, et~al.]{wei2022chain}
J.~Wei, X.~Wang, D.~Schuurmans, M.~Bosma, F.~Xia, E.~Chi, Q.~V. Le, D.~Zhou, et~al.
\newblock Chain-of-thought prompting elicits reasoning in large language models.
\newblock \emph{Advances in neural information processing systems}, 35:\penalty0 24824--24837, 2022.

\bibitem[Wu et~al.(2024)Wu, Hou, Liu, Che, Ju, Yang, Li, Zhao, Xu, Yang, et~al.]{wu2024robomind}
K.~Wu, C.~Hou, J.~Liu, Z.~Che, X.~Ju, Z.~Yang, M.~Li, Y.~Zhao, Z.~Xu, G.~Yang, et~al.
\newblock Robomind: Benchmark on multi-embodiment intelligence normative data for robot manipulation.
\newblock \emph{arXiv preprint arXiv:2412.13877}, 2024.

\bibitem[Xie et~al.(2019)Xie, Xu, Kankanhalli, Meel, and Soh]{xie2019embedding}
Y.~Xie, Z.~Xu, M.~S. Kankanhalli, K.~S. Meel, and H.~Soh.
\newblock Embedding symbolic knowledge into deep networks.
\newblock \emph{Advances in neural information processing systems}, 32, 2019.

\bibitem[Xu et~al.(2025{\natexlab{a}})Xu, Zhang, Guo, Wen, Yang, Lin, Huang, Li, Zhang, Wang, Kuang, Cao, Zheng, and Liang]{a0}
R.~Xu, J.~Zhang, M.~Guo, Y.~Wen, H.~Yang, M.~Lin, J.~Huang, Z.~Li, K.~Zhang, L.~Wang, Y.~Kuang, M.~Cao, F.~Zheng, and X.~Liang.
\newblock A0: An affordance-aware hierarchical model for general robotic manipulation, 2025{\natexlab{a}}.
\newblock URL \url{https://arxiv.org/abs/2504.12636}.

\bibitem[Xu et~al.(2025{\natexlab{b}})Xu, Zhang, Guo, Wen, Yang, Lin, Huang, Li, Zhang, Wang, et~al.]{xu2025a0}
R.~Xu, J.~Zhang, M.~Guo, Y.~Wen, H.~Yang, M.~Lin, J.~Huang, Z.~Li, K.~Zhang, L.~Wang, et~al.
\newblock A0: An affordance-aware hierarchical model for general robotic manipulation.
\newblock \emph{arXiv preprint arXiv:2504.12636}, 2025{\natexlab{b}}.

\bibitem[Yang et~al.(2025{\natexlab{a}})Yang, Tan, Wu, Zheng, Peng, Liang, Gu, Cai, Ye, Jang, et~al.]{magma}
J.~Yang, R.~Tan, Q.~Wu, R.~Zheng, B.~Peng, Y.~Liang, Y.~Gu, M.~Cai, S.~Ye, J.~Jang, et~al.
\newblock Magma: A foundation model for multimodal ai agents.
\newblock \emph{arXiv preprint arXiv:2502.13130}, 2025{\natexlab{a}}.

\bibitem[Yang et~al.(2025{\natexlab{b}})Yang, Li, Chen, Wang, Tian, Wang, Wang, Zhao, Liao, and Pang]{instructvla}
S.~Yang, H.~Li, Y.~Chen, B.~Wang, Y.~Tian, T.~Wang, H.~Wang, F.~Zhao, Y.~Liao, and J.~Pang.
\newblock Instructvla: Vision-language-action instruction tuning from understanding to manipulation.
\newblock \emph{arXiv preprint arXiv:2507.17520}, 2025{\natexlab{b}}.

\bibitem[Ye et~al.(2025)Ye, Jang, Jeon, Joo, Yang, Peng, Mandlekar, Tan, Chao, Lin, Liden, Lee, Gao, Zettlemoyer, Fox, and Seo]{ye2025lapa}
S.~Ye, J.~Jang, B.~Jeon, S.~Joo, J.~Yang, B.~Peng, A.~Mandlekar, R.~Tan, Y.-W. Chao, B.~Y. Lin, L.~Liden, K.~Lee, J.~Gao, L.~Zettlemoyer, D.~Fox, and M.~Seo.
\newblock Latent action pretraining from videos.
\newblock In \emph{The Thirteenth International Conference on Learning Representations (ICLR)}, 2025.

\bibitem[Yu et~al.(2016)Yu, Poirson, Yang, Berg, and Berg]{yu2016modeling}
L.~Yu, P.~Poirson, S.~Yang, A.~C. Berg, and T.~L. Berg.
\newblock Modeling context in referring expressions.
\newblock In \emph{European conference on computer vision}, pages 69--85. Springer, 2016.

\bibitem[Yu et~al.(2023)Yu, Yang, Li, Wang, Lin, Liu, Wang, and Wang]{yu2023mm}
W.~Yu, Z.~Yang, L.~Li, J.~Wang, K.~Lin, Z.~Liu, X.~Wang, and L.~Wang.
\newblock Mm-vet: Evaluating large multimodal models for integrated capabilities.
\newblock \emph{arXiv preprint arXiv:2308.02490}, 2023.

\bibitem[Yuan et~al.(2024)Yuan, Duan, Blukis, Pumacay, Krishna, Murali, Mousavian, and Fox]{yuan2024robopoint}
W.~Yuan, J.~Duan, V.~Blukis, W.~Pumacay, R.~Krishna, A.~Murali, A.~Mousavian, and D.~Fox.
\newblock Robopoint: A vision-language model for spatial affordance prediction for robotics.
\newblock \emph{arXiv preprint arXiv:2406.10721}, 2024.

\bibitem[Zawalski et~al.(2024)Zawalski, Chen, Pertsch, Mees, Finn, and Levine]{ecot}
M.~Zawalski, W.~Chen, K.~Pertsch, O.~Mees, C.~Finn, and S.~Levine.
\newblock Robotic control via embodied chain-of-thought reasoning.
\newblock \emph{arXiv preprint arXiv:2407.08693}, 2024.

\bibitem[Zhai et~al.(2023)Zhai, Mustafa, Kolesnikov, and Beyer]{siglip}
X.~Zhai, B.~Mustafa, A.~Kolesnikov, and L.~Beyer.
\newblock Sigmoid loss for language image pre-training.
\newblock In \emph{Proceedings of the IEEE/CVF International Conference on Computer Vision}, pages 11975--11986, 2023.

\bibitem[Zhang et~al.(2021)Zhang, Zheng, Wu, Fu, and Chen]{zhang2021mme}
Y.~S. Y. Q.~M. Zhang, X.~L. J. Y.~X. Zheng, K.~L. X. S.~Y. Wu, R.~J.~C. Fu, and P.~Chen.
\newblock Mme: A comprehensive evaluation benchmark for multimodal large language models.
\newblock \emph{arXiv preprint arXiv:2306.13394}, 18, 2021.

\bibitem[Zhao et~al.(2025)Zhao, Lu, Kim, Fu, Zhang, Wu, Li, Ma, Han, Finn, et~al.]{zhao2025cot}
Q.~Zhao, Y.~Lu, M.~J. Kim, Z.~Fu, Z.~Zhang, Y.~Wu, Z.~Li, Q.~Ma, S.~Han, C.~Finn, et~al.
\newblock Cot-vla: Visual chain-of-thought reasoning for vision-language-action models.
\newblock In \emph{Proceedings of the Computer Vision and Pattern Recognition Conference}, pages 1702--1713, 2025.

\bibitem[Zhou et~al.(2025{\natexlab{a}})Zhou, An, Chi, Han, Rong, Zhang, Wang, Wang, Huang, Sheng, et~al.]{zhou2025roborefer}
E.~Zhou, J.~An, C.~Chi, Y.~Han, S.~Rong, C.~Zhang, P.~Wang, Z.~Wang, T.~Huang, L.~Sheng, et~al.
\newblock Roborefer: Towards spatial referring with reasoning in vision-language models for robotics.
\newblock \emph{arXiv preprint arXiv:2506.04308}, 2025{\natexlab{a}}.

\bibitem[Zhou et~al.(2025{\natexlab{b}})Zhou, Zhu, Wen, Shen, and Xu]{chatvla2}
Z.~Zhou, Y.~Zhu, J.~Wen, C.~Shen, and Y.~Xu.
\newblock Vision-language-action model with open-world embodied reasoning from pretrained knowledge.
\newblock \emph{arXiv preprint arXiv:2505.21906}, 2025{\natexlab{b}}.

\bibitem[Zhu et~al.(2025)Zhu, Wang, Chen, Liu, Ye, Gu, Tian, Duan, Su, Shao, Gao, Cui, Wang, Cao, Liu, Wei, Zhang, Wang, Xu, Li, Wang, Deng, Li, He, Jiang, Luo, Wang, He, Shi, Zhang, Shao, He, Xiong, Qu, Sun, Jiao, Lv, Wu, Zhang, Deng, Ge, Chen, Wang, Dou, Lu, Zhu, Lu, Lin, Qiao, Dai, and Wang]{internvl3}
J.~Zhu, W.~Wang, Z.~Chen, Z.~Liu, S.~Ye, L.~Gu, H.~Tian, Y.~Duan, W.~Su, J.~Shao, Z.~Gao, E.~Cui, X.~Wang, Y.~Cao, Y.~Liu, X.~Wei, H.~Zhang, H.~Wang, W.~Xu, H.~Li, J.~Wang, N.~Deng, S.~Li, Y.~He, T.~Jiang, J.~Luo, Y.~Wang, C.~He, B.~Shi, X.~Zhang, W.~Shao, J.~He, Y.~Xiong, W.~Qu, P.~Sun, P.~Jiao, H.~Lv, L.~Wu, K.~Zhang, H.~Deng, J.~Ge, K.~Chen, L.~Wang, M.~Dou, L.~Lu, X.~Zhu, T.~Lu, D.~Lin, Y.~Qiao, J.~Dai, and W.~Wang.
\newblock Internvl3: Exploring advanced training and test-time recipes for open-source multimodal models, 2025.
\newblock URL \url{https://arxiv.org/abs/2504.10479}.

\end{thebibliography}

\appendix

\newpage
\section{Author contributions}
\label{sec:authors}
All contributors are listed in \textbf{alphabetical} order by their last names.

\noindent
\subsection{Core Contributors}

\noindent
Yilun Chen, Ning Gao, Jiangmiao Pang, Bolun Wang, Fangjing Wang, Jinhui Ye, Junqiu Yu, Jinyu Zhang, Yangkun Zhu

\noindent
\subsection{Contributors}

\noindent
Xinyi Chen, Yanwei Fu, Jiaya Jia, Weiyang Jin, Hao Li, Yao Mu, Yu Qiao, Yang Tian, Bin Wang, Hanqing Wang, Tai Wang, Ziqin Wang, Xueyuan Wei, Chao Wu, Shuai Yang, Jia Zeng, Jingjing Zhang, Shi Zhang, Feng Zheng, Bowen Zhou

\end{document}